\documentclass[peerreviewca,a4paper,10pt]{IEEEtran} 
\usepackage{amsmath}   

\usepackage{algorithm}
\usepackage{algorithmic}

\usepackage{epsfig}
\usepackage{amsfonts,amssymb}
\usepackage[caption=false]{subfig}

\newtheorem{theorem}{Theorem}
\newtheorem{proposition}{Proposition}
\newtheorem{lemma}{Lemma}
\newtheorem{definition}{Definition}
\newcommand{\msubsubsection}[1]{\vspace{5pt}\subsubsection*{#1}~\vspace{5pt}}
\newcommand{\bdefinition}{\vspace{5pt}\begin{definition}}
\newcommand{\blemma}{\vspace{5pt}\begin{lemma}}
\newcommand{\bproposition}{\vspace{5pt}\begin{proposition}}
\newcommand{\btheorem}{\vspace{5pt}\begin{theorem}}
\newcommand{\bproof}{\vspace{0pt}\begin{IEEEproof}}
\newcommand{\edefinition}{\end{definition}\vspace{5pt}}
\newcommand{\elemma}{\end{lemma}\vspace{5pt}}
\newcommand{\eproposition}{\end{proposition}\vspace{5pt}}
\newcommand{\etheorem}{\end{theorem}\vspace{5pt}}
\newcommand{\eproof}{\end{IEEEproof}\vspace{5pt}}
\begin{document}

\title{On Probabilistic Completeness of Probabilistic Cell Decomposition}

\author{\IEEEauthorblockN{Frank Lingelbach}
\IEEEauthorblockA{Centre for Autonomous Systems\\
KTH Royal Institute of Technology\\
SE-100 44 Stockholm, Sweden\\
Email: frank.lingelbach@ee.kth.se}}

\maketitle

\begin{abstract}
Probabilistic Cell Decomposition (PCD) is a probabilistic path planning method combining the concepts of approximate cell decomposition with probabilistic sampling. It has been shown that the use of lazy evaluation techniques and supervised sampling in important areas result in a high performance path planning method. Even if it was postulated before that PCD is probabilistically complete, we present a detailed proof of probabilistic completeness here for the first time.
\end{abstract}

\section{Introduction}
The problem of path planning appears in many different forms that seem only loosely connected at first. An instance of the problem could, for example, be a point like agent that has to traverse a maze, an articulated robot that has to move from one configuration to another or path planning for a free-flying rigid body. Additional examples include analysis of folds of large molecules, assembly / disassembly planning or path planning for animated characters. They all have in common that some kind of agent has to move from a start to a goal position avoiding collisions with static obstacles and self-collisions along the way. The problem can be transferred to the so called configuration space $\mathcal{C}$ where the problem reduces to finding a continuous path for a point connecting the start with the goal configuration. This comes at the cost that the dimension of $\mathcal{C}$ can be much higher than the dimension of the workspace $\mathcal{W}$. In general, the dimension of $\mathcal{C}$ corresponds to the number of degrees of freedom of the agent. Another drawback of planning in $\mathcal{C}$ is that for all but trivial problems the computation of an explicit representation of obstacles in $\mathcal{C}$ is computationally too expensive to be feasible. Information on $\mathcal{C}$-space obstacles is available through a boolean collision-check function only. An algorithm can probe a specific configuration and gets an answer whether it is collision-free or not. For a given configuration $q$, the collision-check function checks whether the image of $q$ in $\mathcal{W}$ collides with an obstacle in $\mathcal{W}$.

In \cite{lingelbach04a} we introduced a new method for solving path planning queries: Probabilistic Cell Decomposition (PCD). We indicated that PCD is probabilistically complete. However, in this paper, we provide a detailed proof of probabilistic completeness of PCD for the first time. An algorithm is said to be probabilistically complete if for the case that a collision-free path exists, the probability of the algorithm successfully solving the path planning query approaches one when more and more computation time is spent on the problem.

This technical report requires a basic understanding of path planning concepts. A good introduction and a broad overview over the field of path planning are offered by \cite{latombe91,choset05,lavalle06}.

In the following section we briefly introduce the algorithm and its basics components. We will then discuss some challenges in proving probabilistic completeness for PCD compared to other probabilistically complete path planning methods like Probabilistic Roadmaps (PRM) \cite{kavraki96,amato98a,dale01} and Rapidly-exploring Random Trees (RRT) \cite{lavalle98,lavalle00}. Finally, we prove probabilistic completeness for PCD. For further reading on PCD we refer to \cite{lingelbach05a,lingelbach05b, lingelbach06a}.
\section{Probabilistic Cell Decomposition}
\label{Sec:basalg}
PCD resembles an approximate cell decomposition method where cells have a simple predefined shape. As in most approximate cell decomposition methods, PCD divides the configuration space $\mathcal{C}$ into almost disjoint closed rectangloid cells. \emph{Almost} in the sense that neighboring cells share a common boundary and configurations on this boundary belong to both cells.

PCD does not require an explicit representation of the configuration space obstacles but only a binary collision checker that can probe a specific configuration for collision. Thus, it is never known whether a cell is entirely free or entirely occupied by $\mathcal{C}$-obstacles. Instead, a cell is assumed to be free until disproval. A cell is called \emph{possibly free}, as long as only collision-free samples have been found in the cell. Accordingly, it is called \emph{possibly occupied} if all samples that have been checked are colliding. If both collision-free and colliding samples have been found in
the same cell, it is \emph{mixed} and has to be split up into possibly free and possibly occupied cells.

The possibly free cells form the nodes of the connectivity graph $\mathcal{G}$. Two nodes are connected by an edge if the corresponding cells are adjacent. In contrast to classical approximate cell decomposition methods the concept of \emph{possibly} free cells implies unfortunately that from a cell path in $\mathcal{G}$ we cannot automatically deduce an existing feasible continuous path. The states along a continuous path through this cell path have to be checked for collision. Accordingly, from the fact that no cell path exists in $\mathcal{G}$ we cannot conclude that \emph{no} continuous path exists.


\subsection{Notation}
Let us introduce the basic notation.

PCD works in the configuration space $\mathcal{C}$. The open subset of collision-free configurations is denoted by $\mathcal{C}_\textnormal{free}\subseteq\mathcal{C}$ and the set of colliding configurations is given by $\mathcal{C}_\textnormal{obst} = \mathcal{C} \setminus\mathcal{C}_\textnormal{free}$.
PCD will decompose $\mathcal{C}$ into rectangloid cells $\kappa$. Such a cell $\kappa$ can be

\begin{itemize}
    \item possibly free, i.e.
    $P(\kappa\subset\mathcal{C}_\textnormal{free})>0$,
    \item possibly occupied, i.e.
    $P(\kappa\subset\mathcal{C}_\textnormal{obst})>0$,
    \item mixed, i.e.
    $\kappa\not\subset\mathcal{C}_\textnormal{free}\wedge\kappa\not\subset\mathcal{C}_\textnormal{obst}$ or
    $\phantom{blablablablablablablablablablablablablablablablablablablablablablablablablablablablablablablablablablablabla}$
    $\phantom{mixed, i.e}$$P(\kappa\subset\mathcal{C}_\textnormal{free})=P(\kappa\subset\mathcal{C}_\textnormal{obst})=0$.
\end{itemize}

Let $\kappa_\textnormal{start}$ and $\kappa_\textnormal{goal}$ denote the
possibly free cells containing the initial and the goal
configuration, $q_\textnormal{start}$ and $q_\textnormal{goal}$, respectively.
Note that throughout the decomposition process these labels will
be passed to different cells when the respective cells get split.

The possibly free cells form the nodes of the connectivity graph
$\mathcal{G}$. Two nodes are connected by an
edge if and only if the corresponding cells are adjacent. The set
of cells corresponding to the connected component of $\mathcal{G}$
containing $\kappa_\textnormal{start}$ is called the start region
$\mathcal{R}_\textnormal{start}$, accordingly for the goal region
$\mathcal{R}_\textnormal{goal}$. A path in $\mathcal{G}$ connecting
$\kappa_\textnormal{start}$ with $\kappa_\textnormal{goal}$ is called a
\emph{channel} or \emph{cell path} interchangeably.


\subsection{The Basic Algorithm}
Algorithm~\ref{Fig:pcdbalg} shows the basic algorithm of PCD. It is initialized with one possibly free cell $\kappa_\textnormal{start}=\kappa_\textnormal{goal}=\mathcal{C}$. Thus, initially $\mathcal{G}$ has only one single node.
The outer \texttt{while}-loop simply loops forever until a solution has been found. Like other probabilistic path planning methods, PCD is not able to determine that no feasible path exists. One iteration of the outer \texttt{while}-loop is considered \emph{one iteration of PCD}. The inner \texttt{while}-loop (lines 2--9) continuously searches for a cell path connecting $\kappa_\textnormal{start}$ with $\kappa_\textnormal{goal}$. It breaks if no such cell path could be found (\texttt{findCellPath(G) = null}). If a cell path is found, a continuous local path through this channel is checked for collision (\texttt{checkPath(cellPath)}). If the continuous path is found to be collision-free, the path planning query is solved. Otherwise, a collision has been found in a possibly free cell. Thus, this cell is mixed and, in \texttt{splitMixedCells}, all mixed cells are split into possibly free and possibly occupied cells.

When the inner \texttt{while}-loop breaks, $\kappa_\textnormal{start}$ and $\kappa_\textnormal{goal}$ are in different connected components of $\mathcal{G}$. Therefore, the possibly occupied cells get sampled (\texttt{randomState(possiblyOccupiedCells)}) in order to refine them and try to reconnect $\mathcal{R}_\textnormal{start}$ with $\mathcal{R}_\textnormal{goal}$ in $\mathcal{G}$. If one of the samples happens to be collision-free, the corresponding cell is mixed as it contains both colliding and collision-free samples. Thus, this cell has to be split into possibly free and possibly occupied cells.

By alternately sampling and local path checking, the cell decomposition is iteratively refined until a collision-free path is found through a channel of possibly free cells.

The basic algorithm shown in Algorithm \ref{Fig:pcdbalg} is equivalent to the one presented in \cite{lingelbach04a}.

\begin{algorithm}
    \caption{The Basic algorithm of PCD}
    \label{Fig:pcdbalg}
    \begin{algorithmic}[1]
        \WHILE{\NOT success}

            \WHILE{null $\neq$ cellPath $\gets$ findCellPath(G)}
                \IF{checkPath(cellPath)}
                    \STATE success $\gets$ \TRUE
                    \STATE \textbf{break while}
                \ELSE
                    \STATE splitMixedCells
                \ENDIF
            \ENDWHILE
            \IF{\NOT success}
                \STATE samples $\gets$ randomState(possiblyOccupiedCells)
                \IF{\NOT collisionCheck(samples)}
                    \STATE splitMixedCells
                \ENDIF
            \ENDIF
        \ENDWHILE
    \end{algorithmic}
\end{algorithm}

\begin{LaTeXdescription}
  \item[\texttt{findCellPath(G)}.] \texttt{findCellPath} searches the connectivity graph $\mathcal{G}$ given by \texttt{G} for a cell path connecting the start cell $\kappa_\textnormal{start}$ with the goal cell $\kappa_\textnormal{goal}$. In \cite{lingelbach04a} 
      A*-search is used for graph search. For this proof, the actual graph search method is of no particular importance. It just has to be complete in the sense that it finds a path in $\mathcal{G}$ if one exists. If a cell path is found, it is stored in \texttt{cellPath}. Otherwise, \texttt{cellPath} is set to \texttt{null} and the inner \texttt{while}-loop breaks.
  \item[\texttt{checkPath(cellPath)}.] In \texttt{checkPath(cellPath)}, a continuous path through the cell path given by \texttt{cellPath} is checked for collisions. First a continuous local path is derived from the cell path. In \cite{lingelbach04a}
      the strategy is to connect the centers of shared boundaries between adjacent cells along the cell path. Again, the actual choice of local path planning method is not important as long as the local path stays inside the cell it shall traverse and the various local paths together form a continuous path connecting $q_\textnormal{start}$ with $q_\textnormal{goal}$. The continuous path is then checked for collision. We assume here that we have a method that provides an answer in finite time whether the continuous path is collision-free or not. In most implementations this will be done by checking configurations along the path at a given resolution. In contrast to the approach presented in \cite{lingelbach04a}, 
      configurations that have been verified as collision-free in this step are not stored in the respective cell. This considerably simplifies the analysis of the algorithm. We will point out later in the proof where this fact is used.

      If the continuous path is found to be collision-free, \texttt{true} is returned and the path planning query is solved. Otherwise, a collision has been found in a possibly free cell. Thus, this cell is known to be mixed. \texttt{false} is returned and, in the next step, all mixed cells are split into possibly free and possibly occupied cells.
  \item[\texttt{splitMixedCells}.] If a cell contains both colliding and collision-free samples it is mixed and has to be split into possibly occupied and possibly free cells. For this proof we assume that in \texttt{splitMixedCells} all mixed cells are split according to the rules defined in \cite{lingelbach04a}, 
      i.e. cells are split in the middle between two samples of opposing type perpendicular to the dimension of largest distance between these two samples. This guarantees that all cells maintain the rectangloid cell structure.
  \item[\texttt{randomState(pOccCells)}.] \texttt{randomState(pOccCells)} draws one random sample in each possibly occupied cell according to a uniform distribution over this cell. The samples are stored in the vector \texttt{q}.
  \item[\texttt{collisionCheck(q)}.] The samples \texttt{q} drawn on the previous line are then checked for collision in \texttt{collisionCheck(q)}. If all samples are colliding, \texttt{collisionCheck} returns \texttt{true}. Otherwise, if one or more samples in \texttt{q} are collision-free, \texttt{collisionCheck} returns \texttt{false} and the algorithm continues to \texttt{splitMixedCells}.
\end{LaTeXdescription}

The different subroutines are described in more detail in
\cite{lingelbach04a}. Unlike in that paper, we assume that
collision-free samples found in \texttt{checkPath} are not stored
in the corresponding cell. Thereby, we discard information but
simplify the analysis of the algorithm considerably. Only those collision-free
samples are stored that are found by \texttt{randomState} in
possibly occupied cells. Consequently, a possibly free cell
contains only one collision-free sample (apart from the initial
cell containing both the start and goal configuration).

The algorithm maintains two major data structures:

\begin{LaTeXdescription}
    \item[binary tree] The binary tree structure keeps track of all cell splits. The root of the binary tree is the initial cell corresponding to the entire configuration space. When a cell is found to be mixed and split into possibly free and possibly occupied cells, the tree grows deeper from the corresponding leaf. Accordingly, the leafs of the tree are the possibly free and possibly occupied cells. All non-leafs are mixed cells.
    \item[connectivity graph $\mathcal{G}$] The connectivity graph $\mathcal{G}$ holds all possibly free cells as nodes and connects two nodes by an edge iff the two cells are adjacent. Two cells are adjacent iff they share a common boundary with nonzero $(n-1)$-measure.
\end{LaTeXdescription}

Let $k$ count the number of iterations of the outer \texttt{while}-loop. One iteration of the outer \texttt{while}-loop is considered \emph{one iteration of PCD}. Refer to the inner \texttt{while}-loop (lines 2--9) with \texttt{<a>} and count the number of iterations of this inner loop with $k_a$. $k_a$ is reset to zero with every iteration of PCD. Refer to the second \texttt{if}-clause (lines 10--15) with \texttt{<b>}.

\section{Probabilistic Completeness}
\label{Sec:procomp}
In this section we prove that PCD is probabilistically complete.

\begin{definition}{\textbf{Probabilistic Completeness.}}
A path planning algorithm is said to be \emph{probabilistically complete} if, for the case that a feasible
path exists, the probability that the algorithm solves the problem
approaches one as computation time goes to infinity.
\end{definition}
\noindent The biggest drawback of a method which is only probabilistically complete is
the fact that it cannot decide whether a problem is not solvable. If a probabilistically complete method is used on an unsolvable
problem, it will simply run forever. This is in contrast to the much stronger concept of \emph{completeness}.

 One can argue that knowing that an algorithm is probabilistically complete is not particularly useful when deciding which algorithm to choose for a problem at hand. An algorithm that regularly solves complex problems in short time without being probabilistically complete might be a better choice than an algorithm that is proven to find a solution with probability one when time approaches infinity. Unfortunately, \emph{when time goes to infinity} is rarely an option in serious applications. In any case, findings from proving probabilistic completeness help in better understanding the algorithm and its performance. For related methods, results from this kind of analysis have often found their way into improvements of the basic algorithm.

PCD is probabilistically complete and we provide a detailed proof in the following. We will first discuss the challenges in proving probabilistic completeness compared to proving this property for related methods like PRM or RRT. After introducing some required notation we give an outline of the proof followed by an extensive proof in all details.
\subsection{Challenges}

Compared to other probabilistic path planning methods like PRM or RRT, there are some characteristics in PCD that turn out being problematic in proving probabilistic completeness. Proofs of probabilistic completeness of PRM and RRT can be found in \cite{kavraki98}, \cite{ladd04}, \cite{lavalle00} and \cite{kuffner00b}.  All three methods sample the configuration space and build graph structures to represent the connectivity of $\mathcal{C}_\textnormal{free}$. The following listing names the most important differences and how they affect the proof.

\begin{LaTeXdescription}
  \item[Non-uniform sampling.] In the basic PRM or RRT methods, the configuration space is sampled uniformly. In PCD, only the possibly occupied cells get sampled. While this is often speeding up the solution process, one can no longer assume uniform distribution of samples over $\mathcal{C}$ which would be beneficial for proving probabilistic completeness. With the number of samples going to infinity a uniform sampling would be dense with probability one, i.e. any open subset of $\mathcal{C}$ would contain a sample with probability one.
  \item[Cell structure.] The underlying cell structure is instrumental in guiding the sampling. However, it may -- in unfortunate cases -- hinder the path planning process. Even if the existing feasible path has been sampled densely (not in the strict mathematical sense), in PCD a thin possibly occupied cell might block the path. PRM or RRT would identify this path easily.
  \item[Monotonous progress.] In PRM and RRT a node corresponds to a configuration in $\mathcal{C}$ and is added to the connectivity graph (i.e. roadmap or tree) if it is collision-free. An edge between two neighboring nodes in the connectivity graph corresponds to a continuous path between the respective configurations. It is added to the connectivity graph if it is collision-free. The connectivity graph $\mathcal{G}$ is built monotonously and when start and goal configuration fall into the same connected component of $\mathcal{G}$ the path planning query is solved.

      A node in the connectivity graph of PCD is never proven to be free. It represents a cell marked as possibly free since only collision-free samples have been found in this cell so far. An edge in the connectivity graph of PCD does in general not correspond to a specific straight line path in $\mathcal{C}$. In PCD, the local path through a possibly free cell depends on both neighbouring cells along the cell path. If a collision is found while checking this local path, not only an edge is removed from $\mathcal{G}$ but a node. A colliding sample has been found in a -- so far -- possibly free cell. When the node gets removed from $\mathcal{G}$ of course all edges leading to this node have to be removed as well -- even if corresponding local paths were already checked for collision and proven to be collision-free. Hence, there is  -- at first glance -- no monotonous progress in building the connectivity graph.

      It can be shown, however, that even in PCD there is monotonous progress in solving the path planning query.

  \item[Role of colliding samples.] In PRM and RRT colliding samples are not treated further. If a sample is found to be colliding while checking a node or an edge for collision, the respective node or edge is simply not added to the connectivity graph. In PCD colliding samples give rise to possibly occupied cells that may block a channel of possibly free cells. Thus, instead of just looking at the \emph{good} samples that guide us to a continuous path we also have to consider \emph{bad} samples that may actively hinder the solution process.

  \item[Determinism.] In all probabilistic path planning methods there are some parts of the algorithm that are deterministic -- like graph search and connecting neighboring samples with a straight line path -- and some others -- in most cases only the sampling of $\mathcal{C}$ -- that are probabilistic. For the analysis of probabilistic completeness, those deterministic parts are of particular importance that create new samples.

      Taking a look at the basic PRM and RRT algorithms, this does not pose any problem for proving probabilistic completeness. In the basic version of both algorithms, colliding samples are not stored, but only the respective link between two samples is marked as colliding. Collision-free samples found in this step are neither problematic. The proof of probabilistic completeness is solely based on the location of probabilistic samples and the deterministic capability of connecting two adjacent configurations by a straight-line path.

      In PCD deterministically derived samples affect the cell decomposition and hence the whole solution process. One can not assume that all samples in a cell are obtained by probabilistic sampling which would be beneficial for proving probabilistic completeness. Instead, in the following we apply a worst-case-scenario each time deterministic samples are to be considered: where might the deterministic sample pop up in a worst case to hinder the solution process.
\end{LaTeXdescription}

\subsection{Notation}
We will first introduce some more notation required for the proof.

We will prove probabilistic
completeness for the metric space $\mathcal{C}=[0,1]^n\subset\mathbb{R}^n$ with the
Euclidian metric $d$. With $d_i(q,q')=|q_i-q'_i|$ we refer to the pseudometric that gives the distance between two configurations in a single dimension $i$. We use the short notation $d$ or $d_i$ if the two configurations whose distance is to be measured are unambiguous from the given context. With $D_i$ we denote specific distances in dimension $i$ such as $D^\kappa_i$, $D^{B_\varepsilon}_i$ or $D^I_i$ for the width of a cell $\kappa$, of an $\varepsilon$-ball $B_\varepsilon$ or of a further defined intersection $I$.

We will use the Lebesgue measure $\mu^n$ or $\mu$ to measure a \emph{volume} in the $n$-dimensional
configuration space. Accordingly, $\mu^{n-1}$ measures a volume of an $(n-1)$-dimensional subset of $\mathcal{C}$. We will then use fractions of such measures to define probabilities for samples drawn from a uniform distribution.

Let $\mathcal{C}_\textnormal{obst}$ denote the closed semi-algebraic set of configurations from $\mathcal{C}$ where the binary collision checker returns \emph{collision}. Accordingly, $\mathcal{C}_\textnormal{free}=\mathcal{C}/\mathcal{C}_\textnormal{obst}$ is the open collision-free subset of $\mathcal{C}$ and is also semi-algebraic.

\bdefinition[Semi-algebraic sets, \cite{coste2000}]
A semi-algebraic subset of $\mathbb{R}^n$ is the subset of $(x_1,\dots,x_n)$ in $\mathbb{R}^n$ satisfying a boolean combination of polynomial equations and inequalities with real coefficients. In other words, the semi-algebraic subsets of $\mathbb{R}^n$ form the smallest class $\mathcal{SA}_n$ of subsets of $\mathbb{R}^n$ such that:
\begin{enumerate}
  \item If $P\in\mathbb{R}[X_1,\dots,X_n]$, then $\left\{x\in\mathbb{R}^n; P(x)=0\right\}\in\mathcal{SA}_n$ and \\$\left\{x\in\mathbb{R}^n; P(x)>0\right\}\in\mathcal{SA}_n$.
  \item If $A\in\mathcal{SA}_n$ and $B\in\mathcal{SA}_n$, then $A\cup B$, $A\cap B$ and $\mathbb{R}^n\setminus A$ are in $\mathcal{SA}_n$.
\end{enumerate}
\edefinition

Please observe, that the assumption of $\mathcal{C}_\textnormal{obst}$
being a semi-algebraic set is a sufficient but not a necessary
condition for the proof of probabilistic completeness to hold. It is simply a concrete specification of
\emph{$\mathcal{C}_\textnormal{obst}$ has to be sufficiently nice}. On
the other hand, it does not limit the application of this proof with
respect to path planning for mobile manipulation.
Robots and obstacles in $\mathcal{W}$ are usually defined using
semi-algebraic models. These include polygonal and polyhedral
models. Non-semi-algebraic models can be approximated arbitrary closely by
semi-algebraic models. The mappings from $\mathcal{W}$ to
$\mathcal{C}$ used in the common path planning applications, like
translations and rotations, preserve this property. A robot
defined by a semi-algebraic set moving around
$\mathcal{W}$-obstacles defined by semi-algebraic sets yields a
$\mathcal{C}$-obstacle region $\mathcal{C}_\textnormal{obst}$ that can
be described by a semi-algebraic set.

Let $\mathcal{K}$ be the set of all possibly non-mixed cells, i.e. $\mathcal{K}=\mathcal{K}_\textnormal{pfree}\cup\mathcal{K}_\textnormal{pocc}$ where $\mathcal{K}_\textnormal{pfree}$ and $\mathcal{K}_\textnormal{pocc}$ are the sets of all possibly free and possibly occupied cells, respectively.

Let now
$$\gamma : [0,L] \mapsto
\mathcal{C}_\textnormal{free},\quad\gamma(0)=q_\textnormal{start},\quad\gamma(L)=q_\textnormal{goal}$$
denote a continuous path of length $L$ connecting $q_\textnormal{start}$
with $q_\textnormal{goal}$. By $\gamma$ we
also denote the set of configurations corresponding to this path.

Let furthermore $P_\textnormal{Succ}(k)$ be the a priori probability of successfully
finding a path using PCD within $k$ iterations.

\subsection{Outline}

We will first show that without loss of generality we can assume $\gamma$ to be of Manhattan-type and that around any such $\gamma$ there exists a collision-free $\varepsilon$-tunnel with a finite covering by $\varepsilon$-balls. We will then show that
in each iteration of the outer \texttt{while} loop,
\verb"<a>" will break after a bounded number of iterations $k_\textnormal{a}$.
Consequently, the total number of iterations of PCD --- $k$ --- and therefore also the total number of random samples drawn in \texttt{<b>} will go to infinity unless the path planning query is solved before. We will then show that in each iteration of PCD
and for any $\varepsilon$ the probability of finding a sample in an
$\varepsilon$-tunnel around $\gamma$ in \texttt{<b>} is bounded from below.
Thus, the expected number of samples in this tunnel and --- due to the finite covering --- also the expected number of samples in a single $\varepsilon$-ball will go to infinity with
$k$. By showing that for $\varepsilon$ small enough the probability that the number of samples found in a single $\varepsilon$-ball grows unboundedly is zero we will conclude that PCD is probabilistically complete.

\subsection{Proof}
\bproposition
\label{Prop:completeness}
If there exists a continuous path $\gamma$ connecting \emph{$q_\textnormal{start}$} with \emph{$q_\textnormal{goal}$} then
\emph{$$\lim_{k\rightarrow\infty}P_\textnormal{Succ}(k)=1.$$}
\eproposition

In other words: If there exists a continuous feasible path
connecting the start configuration with the goal configuration,
the a priori probability that PCD will have found it after $k$ iterations approaches
one with $k$ going to infinity, i.e., PCD is probabilistically complete.

We start proving probabilistic completeness with some basic observations regarding $\gamma$. We will then step by step follow the outline given above and conclude the correctness of Proposition \ref{Prop:completeness} in the end.

\msubsubsection{We can assume $\gamma$ to be of Manhattan-type. Around $\gamma$ there exists an $\varepsilon$-tunnel with a finite covering by $\varepsilon$-balls}

We will first recall the standard definitions for $\varepsilon$-balls and $\varepsilon$-tunnels. We then show that for any $l\in[0, L]$ around $\gamma(l)$ there exists an $\varepsilon$-ball contained in $\mathcal{C}_\textnormal{free}$ and there exists even an $\varepsilon$ such that the entire $\varepsilon$-tunnel around $\gamma$ is contained in $\mathcal{C}_\textnormal{free}$. From there we show that $\gamma$ can be assumed to be of Manhattan-type --- consisting of sub-paths each parallel to a coordinate axis. For Manhattan-type $\gamma$ we then show that any $\varepsilon$-tunnel has a finite covering by $\varepsilon$-balls.

\bdefinition[$\varepsilon$-ball]
The set $B_n^\infty(\varepsilon,q) = \left\{q'\in\mathcal{C}| \lVert q-q'\rVert_\infty<\varepsilon\right\}$ is called an open $\varepsilon$-ball around $q$. In the following we use the short notation $B_\varepsilon(q)$ for an $n$-dimensional $\varepsilon$-ball and $B^m_\varepsilon(q)$ for any $m$-dimensional $\varepsilon$-ball with $m\le n$ around $q$ according to the $\infty$-norm. For $B_\varepsilon^m(0)$ as frequently used in the definition of sumsets we simply write $B_\varepsilon^m$. Closed $\varepsilon$-balls are obtained by relaxing the strict inequality used to define the corresponding open $\varepsilon$-balls.
\edefinition

Please observe that since we use the $\infty$-norm, the \emph{ball} has the shape of an $n$-dimensional hypercube.

\blemma
\label{Lem:ball} For each collision-free configuration \emph{$q_\textnormal{f}$}
there exists an $\varepsilon>0$ such that all configurations in the
ball \emph{$B_\varepsilon(q_\textnormal{f})$} are collision-free.
\elemma

\bproof
$\mathcal{C}_\textnormal{free}$ is open. By definition of open sets, around any point of an open set there exists an entire $\varepsilon$-neighborhood of points entirely contained in the set.
\eproof

\bdefinition[$\varepsilon$-tunnel]
The set \emph{$T_\varepsilon^\gamma=\bigcup_{l\in[0,L]} B_\varepsilon(\gamma(l)) \subset \mathcal{C}_\textnormal{free}$} is called an $\varepsilon$-tunnel around $\gamma$.
\edefinition

\blemma
\label{Lem:tunnel}
 If there exists a continuous path \emph{$\gamma\subset\mathcal{C}_\textnormal{free}$} connecting \emph{$q_\textnormal{start}$} with \emph{$q_\textnormal{goal}$}, there exists an $\hat{\varepsilon}>0$ with $\forall \varepsilon$ with \emph{$\hat\varepsilon>\varepsilon>0, T^\gamma_\varepsilon\subset\mathcal{C}_\textnormal{free}.$}
\elemma

\bproof
$\gamma$ is in $\mathcal{C}_\textnormal{free}$ which is open. Thus, for every $l\in[0,L]$ there exists an $\varepsilon$ with $B_\varepsilon(\gamma(l))\subset\mathcal{C}_\textnormal{free}$. Since $\gamma$ and $\mathcal{C}_\textnormal{obst}$ are closed and bounded sets, the extreme value theorem states that the continuous function $f(l)=\min_{q_\textnormal{c}\in\mathcal{C}_\textnormal{obst}}(\lVert\gamma(l)-q_\textnormal{c}\rVert_\infty)$ attains its minimum on $l\in[0, L]$. Thus, with $\hat\varepsilon = \min_{l\in[0, L]}f(l)$ and $\hat\varepsilon>\varepsilon>0$ it holds $\forall l \in [0, L],  B_\varepsilon(\gamma(l)) \subset \mathcal{C}_\textnormal{free}$, since $\exists q_\textnormal{c}\in\mathcal{C}_\textnormal{obst}$ with $q_\textnormal{c}\in B_\varepsilon(\gamma(l))$ would imply $\lVert\gamma(l)-q_\textnormal{c}\rVert_\infty < \varepsilon < \hat{\varepsilon}$ which contradicts $\hat{\varepsilon}=\min_{l\in[0,L]}f(l)$. It follows $T_\varepsilon^\gamma=\bigcup_{l\in[0,L]}B_\varepsilon(\gamma(l))\subset\mathcal{C}_\textnormal{free}$.
\eproof

\blemma
\label{Lem:manhattan}
If there exists a continuous path \emph{$\gamma\subset\mathcal{C}_\textnormal{free}$} connecting \emph{$q_\textnormal{start}$} with \emph{$q_\textnormal{goal}$}, there exists a continuous path \emph{$\gamma_\textnormal{Man}\subset\mathcal{C}_\textnormal{free}$} that is of Manhattan type, i.e. it connects \emph{$q_\textnormal{start}$} with \emph{$q_\textnormal{goal}$} by a sequence of sub-paths that are each parallel to a coordinate axis.
\elemma

\bproof
Around any continuous path $\gamma$ there exists a collision-free $\varepsilon$-Tunnel. Now select the following $\varepsilon$-balls $b_i$:

\begin{alignat*}{2}
& & l_1 & =0 \\
&&b_i&=B_\varepsilon(\gamma(l_i)) \\
&&\mathcal{L}_i&=\left\{l|l_i < l\le L, \forall l' \textnormal{ with } l< l'\le L, \gamma(l')\notin b_i\right\} \\
&\textnormal{if }\mathcal{L}_i\neq\emptyset \ \ \ &l_{i+1}&=\min \mathcal{L}_i\\
&\textnormal{else }&l_{i+1} &=L, \textnormal{ stop here}
\end{alignat*}

The first ball is centered at $q_\textnormal{start}$. Each next ball is placed at the configuration where $\gamma$ leaves the current ball. If $\gamma$ contains loops or tight bends and has several fragments intersecting with the current ball, the next ball is placed at the configuration where $\gamma$ leaves the current ball for the last time. If the current ball $b_i$ contains $q_\textnormal{goal}$, $\mathcal{L}_i$ is empty and a final ball is placed at $q_\textnormal{goal}$.

Consecutive balls overlap since for all but the last ball the center of $b_{i+1}$ is a boundary configuration of $b_i$. Obviously, for all but the last ball $l_{i+1}\ge l_i+\varepsilon$. Thus, at most $k_\textnormal{Man}\le L/\varepsilon+1$ $\varepsilon$-balls can be chosen according to this scheme. Now connect the centers of subsequent balls by a shortest Manhattan sub-path $\gamma^i_\textnormal{Man}$ of length $L^i$ with finitely many corners $k^i_\textnormal{cor}$ and $\gamma^i_\textnormal{Man}(0)=\gamma(l_i)$, $\gamma^i_\textnormal{Man}(L^i)=\gamma(l_{i+1})$, $\gamma^i_\textnormal{Man}\subset b_i\cup b_{i+1}$. The concatenation of all these sub-paths gives a Manhattan path $\gamma_\textnormal{Man}\subset\mathcal{C}_\textnormal{free}$ of length $\sum_{i=1}^{k_\textnormal{Man}-1} L^i$ connecting $q_\textnormal{start}$ with $q_\textnormal{goal}$.
\eproof

\begin{figure}
    \centering
    \begin{tabular}{cc}
    \subfloat[$\varepsilon$-tunnel around $\gamma$]{\includegraphics[width=0.30\linewidth]{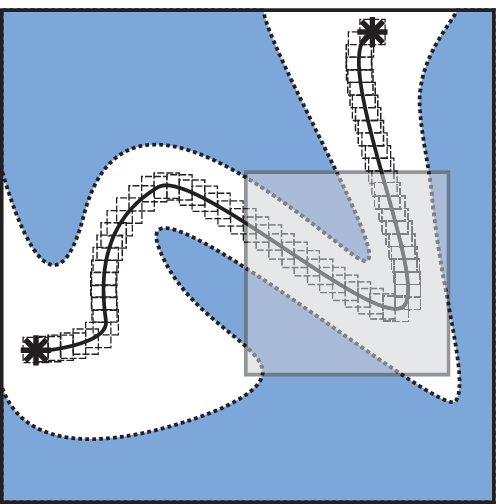}}&
    \subfloat[Enlargement]{\includegraphics[width=0.30\linewidth]{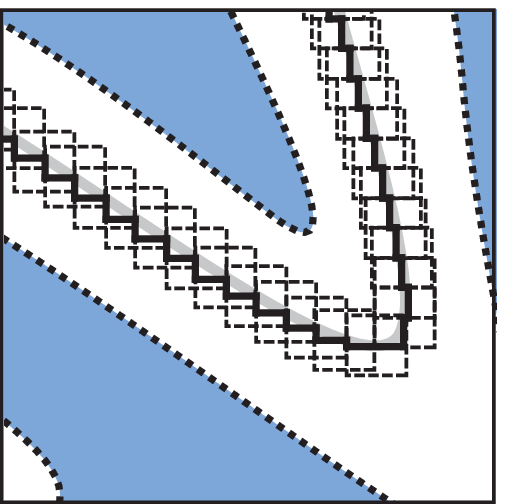}}
    \end{tabular}
    \caption[Existence of Manhattan path]{Configuration space $\mathcal{C}$ with two disjoint obstacle regions (blue, bounded by dotted lines); $\varepsilon$-balls $B_\varepsilon(\gamma(l))$ around selected configurations along the path (dashed squares); (a): start and goal configuration ($\boldsymbol\ast$); continuous feasible path $\gamma$ (solid black line); shaded rectangle shown enlarged in (b); (b): continuous feasible path $\gamma$ (solid gray line); Manhattan path obtained by construction according to Lemma \ref{Lem:manhattan} (solid black line)}
    \label{Fig:epstun}
\end{figure}

As a consequence we can assume without loss of generality that $\gamma$ was of Manhattan type. Figure \ref{Fig:epstun} shows an example of how an existing path $\gamma$ also implies the existence of a Manhattan path $\gamma_\textnormal{Man}$.

\blemma
\label{Lem:covering}
If $\gamma$ is of Manhattan-type, any $T^\gamma_\varepsilon$ can be covered by a finite number \emph{$k_\textnormal{cov}$} of $\varepsilon$-balls, i.e. \emph{$T^\gamma_\varepsilon\subseteq\bigcup_{i=1}^{k_\textnormal{cov}}B_\varepsilon\left(\gamma(l_i)\right)$}.
\elemma

\bproof
For any $\varepsilon>0$ we can construct an $\varepsilon$-tunnel using a finite number of $\varepsilon$-balls, proving equality in the relation above. Place $\varepsilon$-balls at the start and goal configuration and at every corner of the Manhattan path. Then, for any straight-line segment in $\gamma$ that is longer than $\varepsilon$, place $\varepsilon$-balls at intervals of $\varepsilon$. With $k_\textnormal{cor}$ corners in $\gamma_\textnormal{Man}$ it holds $k_\textnormal{cov} < 2+k_\textnormal{cor}+L/\varepsilon$.
\eproof

\msubsubsection{The number of samples in at least one $\varepsilon$-ball around $\gamma$ grows unbounded with the number of iterations of PCD}

We will first show that in each iteration of PCD \texttt{<a>} returns after a bounded number of iterations $k_\textnormal{a}$ and thus the number of iterations of PCD --- $k$ --- goes to infinity. We will then show that in each iteration of PCD at least one possibly occupied cell is sampled that intersects with $\gamma$ and that the probability of finding a sample in an $\varepsilon$-ball around $\gamma$ is bounded from below. Thus the number of samples in an $\varepsilon$-tunnel around $\gamma$ grows unbounded and from the existence of a finite covering of the $\varepsilon$-tunnel by $\varepsilon$-balls we conclude that the number of samples in at least one $\varepsilon$-ball around $\gamma$ grows unbounded with $k$, too.

\bdefinition[Split sectors] The set $$S^i(q)=\left\{q' \in \kappa | d_i(q,q')\ge d_j(q,q') \ \forall j\neq i, 1\le j\le n \right\}$$ is called the $i$-th split sector of a sample $q$ in a cell $\kappa$. The split sector grows from $q$ in both directions ($\pm$). By $S^{i+}(q)=\left\{q' \in S^i(q) | q'_i \ge q_i\right\}$ and $S^{i-}(q)=\left\{q' \in S^i(q) | q'_i \le q_i\right\}$ we denote the \emph{one-sided} $i$-th split sectors.
\edefinition

\begin{figure}
    \centering
    \includegraphics[width=0.30\linewidth]{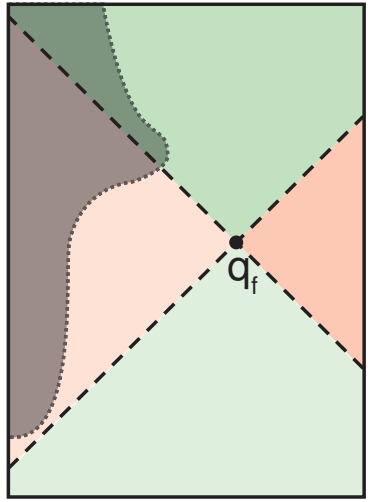}
    \caption[Split sector]{Cell $\kappa$ with collision-free sample $q_\textnormal{f}$ and its split sectors $S^{i-}(q_\textnormal{f})$ (very light red), $S^{i+}(q_\textnormal{f})$ (light red), $S^{j-}(q_\textnormal{f})$ (very light green), and $S^{j+}(q_\textnormal{f})$ (light green) bounded by dashed lines where $i$ and $j$ are the horizontal and vertical dimensions, respectively. A $\mathcal{C}$-space obstacle (shaded darker, bounded by dotted line) intersects with $S^{i-}$ and $S^{j+}$. $S^{i-}(q)=\left\{q' \in \kappa | q'_i \le q_i, d_i(q,q')\ge d_j(q,q') \ \forall j\neq i, 1\le j\le n \right\}$. A colliding sample found in $S^{i-}(q_\textnormal{f})$ leads to a vertical split perpendicular to dimension $i$.}
    \label{Fig:split_sector}
\end{figure}

See Figure~\ref{Fig:split_sector} for an example. Whenever a colliding sample is found in a possibly free cell or a collision-free sample is found in a possibly occupied cell, this cell is marked as mixed and split into possibly free and possibly occupied cells. The cell is split in the middle between the newly found \emph{wrong} sample and the closest old sample. To maintain rectangloid cell shape the cell is split perpendicular to a coordinate axis. In the basic version of PCD the cell is split perpendicular to the dimension of largest distance to the nearest sample of conflictive type. Thus, the cell is split perpendicular to the $i$-th dimension, if the new sample is found in the $i$-th split sector of the nearest old sample. According to the definition above, configurations on the boundary belong to more than one split sector. For this proof, the decision perpendicular to which dimension to split in this case is of no importance.

\blemma \label{Lem:splitbound} In one iteration of PCD, in \verb"<a>" a cell can get split at most
a bounded number of $m$ times. To be more precise, the cell is split into one possibly free and one possibly occupied
cell and the new possibly free cell can get split at most $m-1$ times.
\elemma

\begin{figure}
    \centering
    \includegraphics[width=0.40\linewidth]{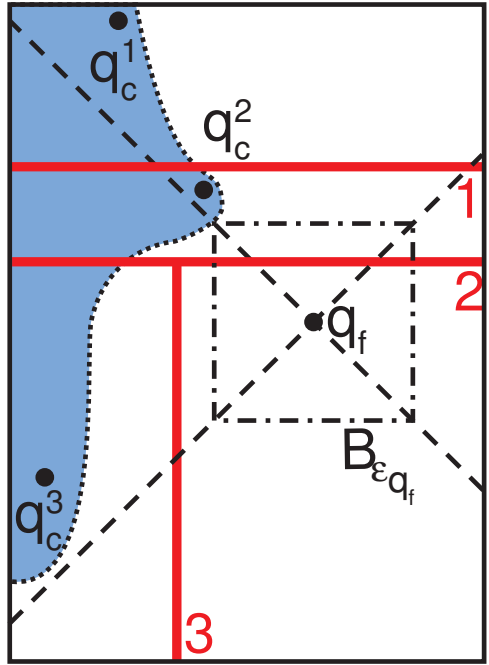}
    \caption[Maximum number of splits in \texttt{<a>}]{Cell $\kappa$ with collision-free sample $q_\textnormal{f}$ and its split sectors (dashed lines). A $\mathcal{C}$-space obstacle (blue, bounded by dotted line) intersects with $S^{i-}$ and $S^{j+}$. After a colliding sample $q^1_\textnormal{c}$ has been found in $S^{j+}(q_\textnormal{f})$ while checking a continuous path in \texttt{<a>} the cell is split horizontally (red line, 1). A second colliding sample $q^2_\textnormal{c}$ found in the next iteration of \texttt{<a>} leads to a second horizontal split (red line, 2). A colliding sample $q^3_\textnormal{c}$ found in $S^{i-}(q_\textnormal{f})$ in the third iteration of \texttt{<a>} leads to a vertical split (red line, 3). After these three splits no more splits are possible since the split sectors of $q_\textnormal{f}$ in the remaining free cell do not contain any colliding configurations. The maximum number of splits possible in one iteration of PCD is determined by the position of $q_\textnormal{f}$ in $\kappa$ and the size of the largest $\varepsilon$-ball $B_{\varepsilon_{q_\textnormal{f}}}$ of collision-free samples surrounding it. }
    \label{Fig:max_splits}
\end{figure}

\bproof
In \verb"<a>" colliding samples may be found while checking a continuous path through a channel of possibly free cells. Each possibly free cell holds exactly one collision-free sample. A colliding sample found in a possibly free cell leads to a cell split. Please recall that a rectangloid cell $\kappa$ is defined by an upper and a lower defining vertex $c^\textnormal{u}_\kappa$ and $c^\textnormal{l}_\kappa$, respectively. We omit the cell index $\kappa$ where the respective cell is nonambiguous to ease notation. Please consult Figure~\ref{Fig:max_splits} for a sketch of the following reasoning.

Let $\varepsilon_{q_\textnormal{f}}$ be the radius of an $\varepsilon$-ball
around the collision-free sample $q_\textnormal{f}\in\kappa$ with
$B_{\varepsilon_{q_\textnormal{f}}}(q_\textnormal{f}) \subset C_\textnormal{free}$. Take any
dimension $i\le n$. The distance in dimension $i$ between the
collision-free sample $q_\textnormal{f}$ and the upper cell boundary $c^\textnormal{u}$ is
denoted by $D_i^\textnormal{u}=d_i(c^\textnormal{u},q_\textnormal{f})$, accordingly for the lower cell
boundary $c^\textnormal{l}$: $D_i^\textnormal{l}=d_i(c^\textnormal{l},q_\textnormal{f})$. If a colliding sample $q_\textnormal{c}$
is found in this cell while checking a continuous path for
collision, the cell has to be split into a possibly free cell and
a possibly occupied cell.

For the split to happen in dimension $i$, $q_\textnormal{c}$ must be found in the $i$-th split sector of $q_\textnormal{f}$, i.e.
it has to hold $d_i(q_\textnormal{f},q_\textnormal{c})=|q_{\textnormal{f}_i}-q_{\textnormal{c}_i}|=\max_jd_j(q_\textnormal{f},q_\textnormal{c})$. A colliding sample $q_\textnormal{c}$ found in $S^{i+}(q_\textnormal{f})$ or $S^{i-}(q_\textnormal{f})$ leads to an \emph{upper split} or a \emph{lower split}, respectively. We present the argument for upper splits only. It holds accordingly for lower splits.

After one
upper split due to a sample $q_\textnormal{c}$ in the $i$-th split sector of $q_\textnormal{f}$ the $i$-th coordinate of the upper boundary of the remaining possibly free cell becomes
$c^{\textnormal{u,new}}_{i}=q_{\textnormal{f}_i}+d_i(q_\textnormal{f},q_\textnormal{c})/2$. The variable with index \emph{old} refers to the value before any split was done according to this lemma, \emph{new} refers to the latest value. Consequently, $D^{\textnormal{u,new}}_{i}=d_i(q_\textnormal{f},q_\textnormal{c})/2\le D^{\textnormal{u,old}}_{i}/2$ and after the
$m$th upper split in direction $i$ it holds $D^{\textnormal{u,new}}_{i}\le
D^{\textnormal{u,old}}_{i}/2^m$. At the latest when
$D^{\textnormal{u,new}}_{i}\le\varepsilon_{q_\textnormal{f}}$ there are no more upper
splits possible as $d_i(q_\textnormal{f},q_\textnormal{c})=\max_jd_j(q_\textnormal{f},q_\textnormal{c})\le
D^{\textnormal{u,new}}_{i}\le\varepsilon_{q_\textnormal{f}}$ contradicts $q\in\mathcal{C}_\textnormal{free}$ for all $q\in B_{\varepsilon_{q_\textnormal{f}}}(q_\textnormal{f})$. Thus, after at
most $m^\textnormal{u}_i=\left\lceil(\log
D^{\textnormal{u,old}}_{i}-\log\varepsilon_{q_\textnormal{f}})/\log2\right\rceil$ upper and
$m^\textnormal{l}_i=\left\lceil(\log
D^{\textnormal{l,old}}_{i}-\log\varepsilon_{q_\textnormal{f}})/\log2\right\rceil$ lower splits there
are no more splits possible in dimension $i$. With $m=\sum_{i=1}^n (m_i^\textnormal{u}
+ m_i^\textnormal{l})$, the lemma holds. After at most $m$ splits, the remaining possibly free cell $\kappa$ is entirely contained in $\mathcal{C}_\textnormal{free}$.
\eproof

\blemma
\label{Lem:a_returns}
In each iteration of PCD, \texttt{<a>} returns after a bounded number of iterations \emph{$k_\textnormal{a}$}.
\elemma
\bproof
This is a direct consequence of Lemma \ref{Lem:splitbound}. In the first iteration of \texttt{<a>} --- $k_a=0$ --- let there be $k_\textnormal{pfree}$ possibly free cells, each containing only one collision-free sample. Collision-free samples found while checking a continuous path for collision in \texttt{<a>} are not stored so the number of collision-free samples stays constant through all iterations of \texttt{<a>} and equals $k_\textnormal{pfree}$. In each iteration of \texttt{<a>} either a continuous path is shown to be collision-free and success is returned or a colliding sample is found and a cell has to be split. Lemma \ref{Lem:splitbound} gives an upper bound on the number of possible cell splits per cell and so the total number of possible cell splits in \texttt{<a>} is bounded by $m_\textnormal{ubound}=\sum_{k=1}^{k_\textnormal{pfree}}m_k$ where $m_k$ is the bound from the preceding lemma for cell $\kappa_k$.

Thus, if the while loop does not break before, after at most
$m_\textnormal{ubound}$ splits all $k_\textnormal{pfree}$ possibly free cells are entirely
contained in $\mathcal{C}_\textnormal{free}$. Consequently, if a cell path can be found, the
continuous path through these cells is collision-free and success
is returned. Otherwise, if no cell path can be found, the while
loop \texttt{<a>} breaks.
\eproof

\blemma
\label{Lem:gamma_sample}
In each iteration of PCD, in \texttt{<b>} at least one cell \emph{$\kappa_\textnormal{pocc}$} is sampled, that contains a fraction of $\gamma$, i.e. there exists a possibly occupied cell \emph{$\kappa_\textnormal{pocc}$} and an $l\in[0,L]$ such that
\emph{$\gamma(l)\in\kappa_\textnormal{pocc}$}.
\elemma

\bproof
In \texttt{<b>} all possibly occupied cells are sampled. Thus we have to show that at least one possibly occupied cell intersects with $\gamma$. Cells are closed sets. The union of all (finitely many) possibly occupied cells $K_\textnormal{pocc}=\bigcup_{\kappa\in\mathcal{K}_\textnormal{pocc}}\kappa$ is hence also closed and the set $A=\mathcal{C}\setminus K_\textnormal{pocc}$ is open.
If for no $l\in[0,L]$ $\gamma(l)$ is contained in a possibly occupied cell (inside or on the boundary) then $\forall l\in [0,L]\ \ \gamma(l)\in A$ which is open and possibly free.
But then there exists an $\varepsilon > 0$ with $T^\gamma_\varepsilon\subset A$.
The continuity of $\gamma$ and thus of $T^\gamma_\varepsilon$ and the strictly positive diameter of $T^\gamma_\varepsilon$ lead to the existence of a series of possibly free cells where consecutive cells are adjacent and share a boundary with nonzero $\mu^{n-1}$. The cells $\phi = \{\kappa|\kappa \in \mathcal{K}_\textnormal{pfree}, \exists l \in [0,L] \gamma(l) \in \kappa\}$ connect the start cell with the goal cell and form a cell path when brought in the right order.
Thus, if there is no such cell path, there has to be an $l\in[0,L]$ and a possibly occupied cell $\kappa_\textnormal{pocc}$ with $\gamma(l)\in\kappa_\textnormal{pocc}$.
\eproof

\blemma
\label{Lem:sample_bound}
In each iteration of PCD, the probability of finding a sample in an $\varepsilon$-tunnel around $\gamma$ is bounded from below.
\elemma

\bproof
It was shown in the previous lemma that when entering the sampling step \texttt{<b>}
at least one cell $\kappa$ along $\gamma$ is possibly occupied.
Since all possibly occupied cells are sampled in \texttt{<b>}, one sample is taken in this cell according to a uniform
distribution. Thus, the probability of finding a sample inside the
$\varepsilon$-tunnel around $\gamma$ in this cell is
$$
P(q\in T^\gamma_{\varepsilon})=\frac{ \mu(\kappa \cap
T^\gamma_{\varepsilon})} {\mu(\kappa)} \geq \frac{\mu(\kappa \cap
B_\varepsilon(\gamma(l)))}{\mu(\kappa)} \textnormal{ for any } l \textnormal{ with } \gamma(l)\in\kappa.
$$
In words: The probability is given by the fraction of the volume of the intersection of the $\varepsilon$-tunnel with the cell and the volume of the whole cell. This fraction is larger or equal to the fraction of the volume of a single $\varepsilon$-ball around $\gamma(l)$ intersected with $\kappa$ and the volume of the cell since the $\varepsilon$-ball is a subset of the $\varepsilon$-tunnel. For the rest of this proof we denote this intersection $\kappa\cap B_\varepsilon(\gamma(l))$ with $I$.

The $n$-dimensional Lebesgue measure
of the rectangloid cell $\kappa$ is given by
$\mu(\kappa)=\prod_{i=1}^n D^\kappa_i$ with $D^\kappa_i=d_i(c^\textnormal{l}, c^\textnormal{u})$. The intersection $I$ of
$\kappa$ with the $\varepsilon$-ball around $\gamma(l)$ is also rectangloid and its measure is given by $\mu(I)=\prod_{i=1}^n D^I_i$, $D^I_i$ defined accordingly with $D^I_i\geq\varepsilon$ if and only if $D^\kappa_i\geq\varepsilon$.
Please recall that $\mathcal{C}=[0, 1]^n$. So, obviously, it holds that
$D^\kappa_i\leq D^\mathcal{C}_i=1$.
But now if
$D^\kappa_i\geq\varepsilon$, then $D^I_i\geq\varepsilon$
and
$\frac{D^I_i}{D^\kappa_i}\geq\frac{\varepsilon}{D^\mathcal{C}_i}$.
If $D^\kappa_i<\varepsilon$, then $D^I_i=D^\kappa_i$
and thus
$\frac{D^I_i}{D^\kappa_i}=1\geq\frac{\varepsilon}{D^\mathcal{C}_i}$.
Consequently,
$$
P(q\in T^\gamma_{\varepsilon}) \geq
\frac{\mu(I)}{\mu(\kappa)}=\frac{\prod_{i=1}^n D^I_i}{\prod_{i=1}^n
D^\kappa_i}\geq\frac{\varepsilon^n}{\mu(\mathcal{C})}=\varepsilon^n
$$
which is independent of the actual cell $\kappa$ and therefore
suitable as a lower bound.
\eproof

\begin{figure}
    \center
    \begin{tabular}{c}
    \epsfig{file=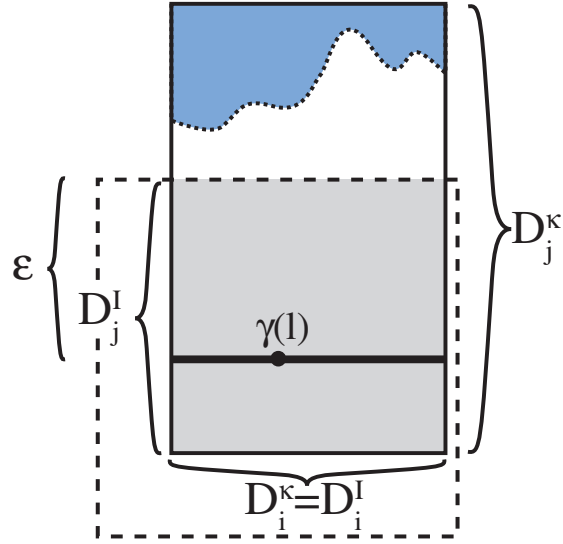, width=0.4\linewidth}
    \end{tabular}
    \caption[Possibly occupied cell intersecting with an $\varepsilon$-ball]
    {Possibly occupied cell (black box) intersecting (gray-shaded box) with an $\varepsilon$-ball (dashed box) around a configuration on the path (bold line) $\gamma(l)$ (black circle); $\mathcal{C}_\textnormal{obst}$: blue, bounded by dotted line}
    \label{Fig:intersec}
\end{figure}

Figure \ref{Fig:intersec} shows a 2D example. A possibly occupied
cell $\kappa$ contains a segment of $\gamma$. The probability of finding a
sample in $T^\gamma_\varepsilon$ is larger or equal to the probability of
finding a sample in a single $\varepsilon$-ball around $\gamma(l)$ for $\gamma(l)\in\kappa$. When sampling the
possibly occupied cell, this probability is equal to the fraction
of the measure of the intersection of the ball with the cell and
the measure of the cell. As shown above, this fraction is bounded
from below and, consequentially, the probability of finding a
sample in the $\varepsilon$-tunnel is bounded from below.

\blemma
\label{Lem:sample_bound_tight}
For a specific cell $\kappa$ the lower bound on finding a sample in an $\epsilon$-tunnel around $\gamma$ derived in Lemma~\ref{Lem:sample_bound} can be increased to
$$
P(q\in T^\gamma_{\varepsilon}) \geq \varepsilon^{n-k_\textnormal{wid}}
$$
where $k_\textnormal{wid}$ is the number of dimensions in which the cell has a width of $D^\kappa_i<\varepsilon$.
\elemma
\bproof
This follows directly from
$$
P(q\in T^\gamma_{\varepsilon}) \geq
\frac{\mu(I)}{\mu(\kappa)}=\frac{\prod_{i=1}^n D^I_i}{\prod_{i=1}^n
D^\kappa_i}
$$
and from the reasoning shown in the proof of the preceding lemma together with the fact that $D^\kappa_i=D^I_i$ for $D^\kappa_i<\varepsilon$.
\eproof

\blemma
\label{Lem:tunnel_samples}
With the number of iterations of PCD going to infinity, the expected number of samples in any $\varepsilon$-tunnel grows unbounded.
\elemma

\bproof
According to Lemma~\ref{Lem:gamma_sample}, in every iteration of PCD at least one possibly occupied cell is sampled that intersects with $\gamma$ and hence intersects with $T^\gamma_\varepsilon$. Lemma~\ref{Lem:sample_bound} showed that the probability of finding a sample inside $T^\gamma_\varepsilon$ is bounded from below. Therefore, with the stochastic variable $\#^\varepsilon_k$ being the number of samples found in the $\varepsilon$-tunnel after $k$
iterations of PCD, its expected value is bounded from below by
$$
E(\#^\varepsilon_k)\geq k\varepsilon^n
$$
which goes to infinity with $k$.
\eproof

\blemma
\label{Lem:single_ball}
With the number of iterations of PCD going to infinity, the expected number of samples inside at least one $\varepsilon$-ball around $\gamma$ grows unbounded.
\elemma

\bproof
Lemma~\ref{Lem:covering} states that any $T^\gamma_\varepsilon$ can be covered by a finite number $k_\textnormal{cov}$ of $\varepsilon$-balls. Thus, in at least one $\varepsilon$-ball the expected number of samples is larger or equal to the expected number of samples in $T^\gamma_\varepsilon$ divided by $k_\textnormal{cov}$. According to Lemma \ref{Lem:tunnel_samples} the expected number of samples in $T^\gamma_\varepsilon$ grows unbounded. Therefore, in at least one of the $\varepsilon$-balls of the finite covering of $T^\gamma_\varepsilon$ the expected number of samples grows unbounded.
\eproof

\subsection*{The probability that the number of samples inside an $\varepsilon$-ball goes to infinity is zero}

We have shown so far that if the path planning query is not solved before and $k$ goes to infinity, the expected number of samples in at least one $\varepsilon$-ball grows unboundedly. We will show in the following that the probability of this to happen is zero. We will first prove that around any collision-free sample there exists a neighborhood that will never again be contained in a possibly occupied cell. The neighborhood is \emph{cleared}. We will then define a subset of an $\varepsilon$-ball --- i.e. its intersection with cells holding a fraction of the path --- and show that even in at least one of these \emph{pruned} $\varepsilon$-balls the expected number of samples would grow unboundedly with $k$ if the path planning query does not get solved before. For these pruned $\varepsilon$-balls we will then show that if there exists a continuous path $\gamma$, the probability of finding an unboundedly growing number of samples in the pruned ball is zero.
We will prove that when more and more samples are found in a pruned $\varepsilon$-ball around $\gamma$, the probability that the entire pruned $\varepsilon$-ball gets cleared around $\gamma$ approaches 1. At the latest when all pruned $\varepsilon$-balls of the finite covering of a pruned $T^\gamma_\varepsilon$ are cleared, the path planning query is solved.

\bdefinition
A collision-free configuration is called \emph{cleared} if it will never
again be contained in a possibly occupied cell. A subset \emph{$S\subset\mathcal{C}_\textnormal{free}$} is called cleared if its interior will never again be contained in a possibly occupied cell. Accordingly, a cell $\kappa$ is called cleared if its interior will never again be contained in a possibly occupied cell. A cell is called \emph{cleared around $\gamma$ for $\varepsilon$} if the interior of its intersection with $T^\gamma_\varepsilon$ is cleared.
\edefinition

The limitation to the interior of sets is necessary to ease notation later on because configurations on the boundary between cells belong to both cells and can be cleared in one cell but not in the other. If all cells intersecting with such a boundary configuration are cleared also the boundary configuration is cleared.

\begin{figure}
    \center
    \begin{tabular}{c}
    \epsfig{file=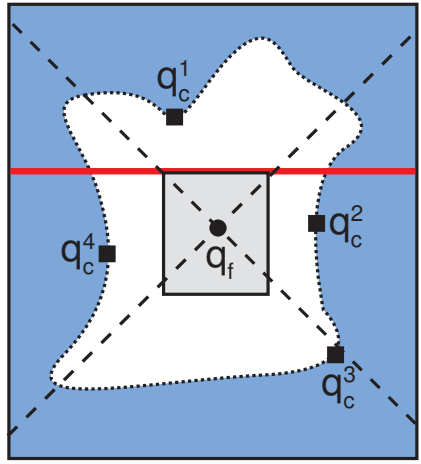, width=0.3\linewidth}
    \end{tabular}
    \caption[Cleared neighborhood around collision-free sample]{Possibly free cell with a collision-free sample $q_\textnormal{f}$ (black circle); $\mathcal{C}_\textnormal{obst}$ (blue, bounded by dotted line); split sectors (dashed lines); closest colliding configuration in each sector (black squares); open neighborhood that will never again be contained in a possibly occupied cell (grey box). If a colliding sample would be found at $q^1_\textnormal{c}$ the cell would be split at the red line. Accordingly for the closest colliding configurations in the other split sectors. The closest possible splits define the \emph{cleared} region in the center that will never again be contained in a possibly occupied cell.}
    \label{Fig:closplit}
\end{figure}

\blemma
\label{Lem:cleared_neighborhood}
Around any collision-free sample there exists an open neighborhood that is cleared, i.e. it will never again be contained in a possibly occupied cell.
\elemma
\bproof
A collision-free sample must be contained in a possibly free cell. A possibly free cell is split only if a colliding sample is found in this cell. As shown in the proof for Lemma~\ref{Lem:splitbound}, in any dimension the closest colliding configuration in the respective split sector determines the closest possible split in this dimension. See Figure~\ref{Fig:closplit} for an example of a collision-free sample $q_\textnormal{f}$ and the closest possible splits determining a neighborhood around $q_\textnormal{f}$ that is cleared. Here it is important, that we store only one collision-free sample per possibly free cell. If there were more, a split between one collision-free sample and a colliding sample could be arbitrarily close to another collision-free sample. All configurations within the rectangloid defined by the closest possible splits according to the respective split sectors will never again be contained in a possibly occupied cell.
\eproof

\blemma
\label{Lem:cleared_ball}
There exists an $\hat{\varepsilon}>0$ such that for all $\hat{\varepsilon}\ge\varepsilon>0$ a sample found inside an $\varepsilon$-ball around $\gamma$ clears the entire intersection of the $\varepsilon$-ball and the respective cell that was sampled.
\elemma

\bproof
 Take $\tilde{\varepsilon} = \sup\left\{\varepsilon\ |\ T_{\varepsilon}^\gamma\in\mathcal{C}_\textnormal{free}\right\}$ as the maximum $\tilde{\varepsilon}$ such that for all $\varepsilon<\tilde{\varepsilon}$ the $\varepsilon$-tunnel around $\gamma$ is still entirely contained in $\mathcal{C}_\textnormal{free}$ and take $\hat\varepsilon=\tilde\varepsilon/5$. Then for any $B_\varepsilon(\gamma(l))$ with $\hat\varepsilon\ge\varepsilon>0$ it holds: If a collision-free sample $q_\textnormal{f}\in B_\varepsilon(\gamma(l))$ is found, then in any split sector of $q_\textnormal{f}$ and any direction ($\pm$) the distance between the collision-free sample $q_\textnormal{f}$ and the closest colliding sample $q_\textnormal{c}$ is not less than $4\varepsilon$ and thus no less than the double distance between $q_\textnormal{f}$ and the boundary of the $\varepsilon$-ball in this direction. But then the closest possible split cannot cut through the $\varepsilon$-ball and hence all configurations within the intersection of $B_\varepsilon(\gamma(l))$ with the respective cell will never again be contained in a possibly occupied cell.
\eproof

\begin{figure}
    \center
    \begin{tabular}{c}
    \epsfig{file=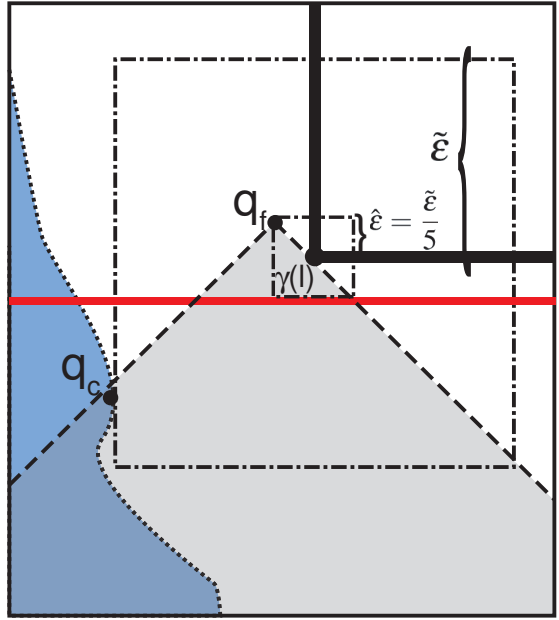, width=0.4\linewidth}
    \end{tabular}
    \caption[Cleared set]{Possibly free cell $\kappa$ with a collision-free sample $q_\textnormal{f}$ (black circle) found in an $\hat{\varepsilon}$-ball (small dash-dotted rectangle) around $\gamma$ (bold black line); $\mathcal{C}_\textnormal{obst}$ (blue, bounded by dotted line); relevant split sector (shaded gray bounded by dashed lines);
    According to Lemma~\ref{Lem:cleared_ball}, if there exists an $\tilde{\varepsilon}$-tunnel around $\gamma$, then any collision-free sample found in an $\hat{\varepsilon}$-ball around $\gamma$ with $\hat{\varepsilon}=\tilde{\varepsilon}/5$ clears the entire intersection of the $\hat{\varepsilon}$-ball with $\kappa$. If then a colliding sample $q_\textnormal{c}$ would be found just outside the $\tilde{\varepsilon}$-ball (large dash-dotted rectangle) as shown in the figure, the closest possible split (red line) would lie just outside the $\hat{\varepsilon}$-ball. Thus, the entire $\hat{\varepsilon}$-ball is cleared.
    }
    \label{Fig:eps5eps}
\end{figure}

In the following we assume $\varepsilon < \tilde\varepsilon/5$ for any $\varepsilon$-ball around $\gamma$ and keep in mind $\hat{\varepsilon} = \tilde{\varepsilon}/5$ as the upper bound in terms of this lemma. Figure~\ref{Fig:eps5eps} shows an example for a collision-free sample $q_\textnormal{f}$ found in the upper left corner of an $\hat{\varepsilon}$-ball around $\gamma$. A colliding sample $q_\textnormal{c}$ is found in the downward pointing split sector of $q_\textnormal{f}$. Hence, the cell is to be split horizontally in the middle between $q_\textnormal{f}$ and $q_\textnormal{c}$. From $\hat{\varepsilon} = \tilde{\varepsilon}/5$ it is guaranteed that any colliding configuration has a minimum distance to $q_\textnormal{f}$ of at least $4\hat{\varepsilon}$ in any dimension. Thus, it is guaranteed that the split will not cut through the $\hat{\varepsilon}$-ball.

\blemma
\label{Lem:straight_or_corner}
There exists an $\overline{\varepsilon}$ such that for all $\overline{\varepsilon}>\varepsilon>0$ and $\forall l\in[0, L]$, $B_\varepsilon(\gamma(l))$ contains either a single straight line segment of the path or a single corner connecting two straight line segments.
\elemma

\bproof
From the construction of $\gamma_\textnormal{Man}$ it has finite length $L$, finitely many corners $k_\textnormal{cor}$ and no loops or tight bends. But then it consists of also finitely many straight line segments. Let the shortest one have length $L^\textnormal{min}>0$ such that each sub-path between two corners is a straight line segment of length $L_i\ge L^\textnormal{min}$. Then any $\varepsilon$-ball $B_\varepsilon(\gamma(l))$ with $L^\textnormal{min}/2=\overline{\varepsilon}>\varepsilon>0$ contains either a straight line segment or a corner connecting two straight line segments and the fraction of $\gamma$ that intersects with $B_\varepsilon(\gamma(l))$ is not greater than $2\varepsilon$.
\eproof
In the following we assume $\varepsilon < \overline{\varepsilon}$ for $\varepsilon$-balls around $\gamma$.

We now would like to show that all $\varepsilon$-balls of the finite covering of the $\varepsilon$-tunnel get cleared when the number of iterations of PCD goes to infinity and conclude that PCD is probabilistically complete. However, we have shown so far that at least one cell intersecting with $\gamma$ is sampled in each iteration of PCD --- instead of one cell intersecting with an $\varepsilon$-ball or $\varepsilon$-tunnel around $\gamma$. For these cells we have then bounded the probability of finding a sample inside the $\varepsilon$-tunnel. So we have no bounds on probabilities for samples in cells that intersect with the $\varepsilon$-ball around $\gamma$ but not with $\gamma$ itself. For the following analysis we therefore have to draw on \emph{pruned} $\varepsilon$-balls and \emph{pruned} $\varepsilon$-tunnels as defined below.

\begin{figure}
    \center
    \begin{tabular}{c}
    \epsfig{file=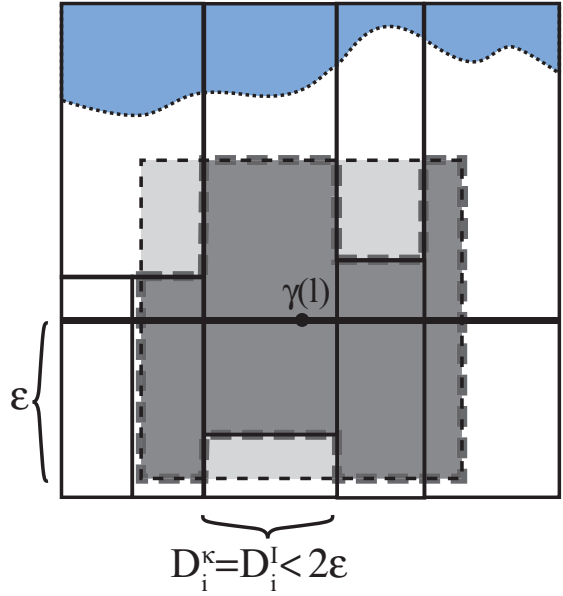, width=0.4\linewidth}
    \end{tabular}
    \caption[Pruned $\varepsilon$-ball]
    {$\varepsilon$-ball (thin dashed line shaded light gray) intersecting with several cells (black contours); the pruned $\varepsilon$-ball $\overline{B}_\varepsilon$ (bold dark gray dashed line shaded gray) is given by the intersection of the $\varepsilon$-ball $B_\varepsilon(\gamma(l))$ with all cells intersecting with the path $\gamma$ (bold line) in $B_\varepsilon$; $\mathcal{C}_\textnormal{obst}$: blue, bounded by dotted line}
    \label{Fig:pruned_ball}
\end{figure}

\bdefinition[Pruned $\varepsilon$-ball]
The set $B_\varepsilon(\gamma(l))\cap\bigcup_{\kappa\in\mathcal{K},\kappa\cap\gamma\neq\emptyset}\kappa$ is called the \emph{pruned $\varepsilon$-ball} around $\gamma$ and denoted by $\overline{B}_\varepsilon(\gamma(l))$. It is the intersection of a standard $\varepsilon$-ball with the union of all cells intersecting with $\gamma$. See Figure \ref{Fig:pruned_ball} for an example.
\edefinition

\bdefinition[Pruned $\varepsilon$-tunnel]
Accordingly, the \emph{pruned $\varepsilon$-tunnel} $\overline{T}_\varepsilon^\gamma$ is obtained by the union of all $\overline{B}_\varepsilon$ for $B_\varepsilon$ in the finite covering of $T_\varepsilon^\gamma$.
\edefinition

\bdefinition[Inner cells, lower outer and upper outer cells]
A cell $\kappa$ intersecting with $\gamma$ in a pruned $\varepsilon$-ball $b$ is called an \emph{inner cell} of $b$ if for $l_\textnormal{min}=\min\{l|l\in[0,L],\gamma(l)\in\kappa\}$ and $l_\textnormal{max}=\max\{l|l\in[0,L],\gamma(l)\in\kappa\}$ it holds $\gamma(l_\textnormal{min})\in b$ and $\gamma(l_\textnormal{max})\in b$. If either $\gamma(l_\textnormal{min})\not\in b$ or $\gamma(l_\textnormal{max})\not\in b$, $\kappa$ is called an \emph{outer cell} of $b$. For being able to be precise later in the proof, we have to distinguish between \emph{lower} and \emph{upper} outer cells. An outer cell $\kappa$ is called \emph{lower} outer cell if $\gamma$ leaves $b$ in $\kappa$ through $c^{b\textnormal{l}}_i$ or \emph{upper} if it leaves $b$ through $c^{b\textnormal{u}}_i$. The case that $\kappa$ spans the ball and neither $l_\textnormal{min}$ nor $l_\textnormal{max}$ is in $b$ is not relevant for the following.
\edefinition

For example, the pruned $\varepsilon$-ball $b$ in Figure \ref{Fig:pruned_ball} has two inner cells and two outer cells, one lower outer cell at the left edge and one upper outer cell at the right.

\blemma
\label{Lem:holds_also_pruned}
Lemmas~\ref{Lem:cleared_ball} and \ref{Lem:straight_or_corner} hold also for pruned $\varepsilon$-balls. For $\varepsilon<\hat{\varepsilon}$ according to Lemma \ref{Lem:cleared_ball}, Lemmas \ref{Lem:sample_bound} -- \ref{Lem:single_ball} hold also for pruned $\varepsilon$-balls and $\varepsilon$-tunnels.
\elemma

\bproof
For Lemma \ref{Lem:cleared_ball}: The intersection of the $\varepsilon$-ball and the cell that was sampled equals the intersection of the pruned $\varepsilon$-ball and this cell. Thus, the same argument holds also for pruned $\varepsilon$-balls.

For Lemma \ref{Lem:straight_or_corner}: Since by definition of pruned $\varepsilon$-balls $\overline{B}_\varepsilon(\gamma(l))\cap\gamma=B_\varepsilon(\gamma(l))\cap\gamma$, the Lemma holds also for pruned $\varepsilon$-balls.

For Lemma \ref{Lem:sample_bound}: Again, the lower bound was derived from a cell $\kappa$ intersecting with $\gamma$ and an $\varepsilon$-ball $B_\varepsilon(\gamma(l))$ with $\gamma(l)\in\kappa$. Thus, the result holds also as lower bound for the pruned $\varepsilon$-ball. Accordingly for Lemma~\ref{Lem:sample_bound_tight}.

For Lemma \ref{Lem:tunnel_samples}: First we notice that the volume of a pruned $\varepsilon$-ball might decrease when a cell is split and one of the two new cells no longer intersects with $\gamma$. However, for $\varepsilon<\hat{\varepsilon}$ the number of samples in this pruned $\varepsilon$-ball will never decrease since a sample $q\in \kappa \cap \overline{B}_\varepsilon(\gamma(l))$ clears the entire intersection of the ball with the cell. Thus there cannot be a split through this intersection with one new cell possibly not intersecting with $\gamma$. Since the probability of finding a sample in a pruned $\varepsilon$ ball in any iteration is bounded from below and since $q \in \overline{B}_\varepsilon(\gamma(l))$ for some $l\in [0,L] \Rightarrow q\in\overline{T}_\varepsilon^\gamma$, the expected number of samples in any pruned $\varepsilon$-tunnel grows also unbounded.

For Lemma \ref{Lem:single_ball}: From the construction of the pruned $\varepsilon$-tunnel it has a finite covering by pruned $\varepsilon$-balls and consequentially the expected number of samples in at least one pruned $\varepsilon$-ball grows unbounded.
\eproof

Consequently, it is sufficient to show that neither a pruned $\varepsilon$-ball containing a straight line segment nor a pruned $\varepsilon$-ball containing a corner of $\gamma$ can be sampled arbitrarily often without clearing the pruned $\varepsilon$-ball. For both cases we will show that when the number of samples in such a pruned $\varepsilon$-ball grows, the probability that the entire ball is cleared approaches one.

We will first take a look at cells and pruned $\varepsilon$-balls that contain a straight line segment of $\gamma$. Later, we will extend these results also to pruned $\varepsilon$-balls containing a corner of $\gamma$.

\bdefinition[Traversing dimension]
For a cell or pruned $\varepsilon$-ball intersecting with a straight line segment of the Manhattan path only, we call the dimension whose axis runs parallel to the path segment the \emph{traversing dimension}.
\edefinition

Since, according to Lemmas \ref{Lem:cleared_ball} and \ref{Lem:holds_also_pruned}, a sample inside a pruned $\varepsilon$-ball $\overline{B}$ clears the entire intersection of the ball and the respective cell, the only way for the ball to contain more than one sample is to intersect with several cells. At any time there also has to be at least one cell intersecting with the ball that has not yet fetched a sample in $\overline{B}$. Consequentially, if the number of samples is to go to infinity, the number of cells it intersects with has to grow unbounded to. The number of cells that intersect with the pruned ball can only grow, if a cell that already intersects with the $\varepsilon$-ball is split in the traversing dimension $i$. A split in any other dimension leads to two cells, one of which does not intersect with $\gamma$ and has to be excluded from the pruned $\varepsilon$-ball. Thus, after a split in any other dimension than $i$, the number of cells intersecting with the pruned $\varepsilon$-ball stays constant.

In any of these cells that already contain a sample in the pruned $\varepsilon$-ball the entire intersection of the pruned $\varepsilon$-ball with the cell is cleared. Hence, for the ball to be able to be sampled again, there have to be cells that have not been cleared yet that intersect with the pruned $\varepsilon$-ball. For the number of samples to grow unbounded, the number of potentially to be sampled cells may never decrease to zero. Once a sample has been found in each cell in $\overline{B}$ the entire pruned ball is cleared.

We will show that there exists an $\varepsilon>0$ such that when a cell intersecting with the pruned $\varepsilon$-ball is sampled, the probability that the cell is split in the traversing dimension is less than the probability that the intersection of the pruned $\varepsilon$-ball with the cell is cleared.

We will now investigate what might happen if such a cell is sampled. The sample found can either be colliding or collision-free. To be sampled in \texttt{<b>} the cell obviously has to be possibly occupied.
\begin{LaTeXdescription}
  \item[colliding] If the sample is colliding, the cell is still possibly occupied. Even though we \emph{know} that it intersects with $\gamma$ and cannot be entirely occupied, the algorithm has found only colliding samples inside so far. So for PCD it is still possibly (entirely) occupied. We now just have another colliding sample in this cell. Since we do not assume to know anything about the other colliding samples found before, we do not care about this sample either. Since the sample found in a possibly occupied cell is colliding, no cell is split and the number of cells that could possibly fetch a sample in the pruned $\varepsilon$-ball is not affected.
  \item[collision-free] If the sample is collision-free, $\kappa$ is mixed and has to be split into possibly occupied and possibly free cells. Since we do not assume to know anything about the location of colliding samples, we investigate the possible splits according to the location of the collision-free sample. To bound the number of splits in the traversing dimension, we assume colliding samples to be found in the worst possible places.
\end{LaTeXdescription}

Please recall that \texttt{<a>} is completely deterministic and produces colliding samples only. The collision-free samples found in \texttt{<a>} while checking a continuous path through possibly free cells are not stored and will not have any impact on the further solution process. Since \texttt{<a>} is deterministic we cannot assume a uniform or any other stochastic distribution of the colliding samples generated in \texttt{<a>}. We therefore show that even in a worst case where colliding samples are found at the most unfavorable locations, the number of samples in an $\varepsilon$-ball cannot grow unbounded.

A collision-free sample found in a cell intersecting with a pruned $\varepsilon$-ball around $\gamma$ in \verb"<b>" leads to one of the following splits:
\begin{LaTeXdescription}
  \item[split in traversing dimension] A split in the traversing dimension gives rise to two cells both intersecting with $\gamma$ and hence with the pruned $\varepsilon$-ball. Only splits in the traversing dimension give rise to more cells intersecting with the pruned $\varepsilon$-ball. Thus, if the number of splits in the traversing dimension can be bounded, we can also bound the number of samples in the pruned $\varepsilon$-ball.
  \item[split in non-traversing dimension] If the collision-free sample leads to a split in a non-traversing dimension, $\gamma$ will after the split either intersect with one possibly free cell or one possibly occupied cell. The other cell arising from this split does not intersect with $\gamma$ in the $\varepsilon$-ball and its interior does hence not intersect with the pruned $\varepsilon$-ball. The number of cells intersecting with $\gamma$ and hence with the pruned $\varepsilon$-ball stays constant.
  \item[clearing split] A special form of the non-traversing split is the \emph{clearing split} that arises from a sample found within an $\hat{\varepsilon}$-ball around $\gamma$. The entire intersection of the $\hat{\varepsilon}$-ball with $\kappa$ is cleared, i.e. the new possibly free cell is cleared around $\gamma$ and will never again be split through the $\hat{\varepsilon}$-ball -- especially not in the traversing dimension. The remaining possibly occupied cell does not intersect with $\gamma$ in the $\varepsilon$-ball and its interior does hence not intersect with the pruned $\varepsilon$-ball. The number of cells that could possibly fetch a sample in the pruned $\varepsilon$-ball in future decreases by one.
\end{LaTeXdescription}

Here, we can identify a monotonous progress in solving the path planning query: more and more fractions of the path get cleared and will never again be contained in a possibly occupied cell. If PCD could not find a collision-free continuous path before, at the latest when all fractions are cleared, the path planning query is solved.

We will now identify probabilities for the events above. In particular, we will bound the probability of a \emph{good} sample -- clearing the entire intersection of the pruned ball with the cell -- from below. Then we bound the probability of a \emph{bad} sample -- potentially leading to a split in the traversing dimension through the pruned ball -- from above. By showing that this upper bound goes to zero with $\varepsilon\rightarrow0$ while the lower bound does not decline, we conclude that there is an $\varepsilon$ where the probability of a good sample exceeds the probability of a bad sample.

\msubsubsection{There exists an $\varepsilon$ such that the probability that the number of samples in a pruned $\varepsilon$-ball goes to infinity is zero.}

We have shown in Lemma~\ref{Lem:straight_or_corner} that for $\varepsilon$ small enough a pruned $\varepsilon$-ball holds either a single corner or a straight line segment of the path $\gamma$. The number of corners of $\gamma$ $k_\textnormal{cor}$ returned from the algorithm presented in Lemma~\ref{Lem:manhattan} is arbitrary but fixed and independent of the choice of $\varepsilon$ for the following analysis of $\varepsilon$-balls. Thus, when we let $\varepsilon$ go to zero, the number of pruned $\varepsilon$-balls from the finite covering of $\gamma$ that hold a single corner stays constant while the number of pruned $\varepsilon$-balls that hold a straight line segment of $\gamma$ increases.

Let $\varepsilon$ be smaller than $\hat{\varepsilon}$ of Lemma~\ref{Lem:cleared_ball}. Thus, a sample found in the pruned $\varepsilon$-ball clears the entire intersection of the sampled cell with the pruned $\varepsilon$-ball.

We start with showing that there exists an $\varepsilon$ such that when an inner cell of a pruned $\varepsilon$-ball is sampled, the probability of finding a \emph{good} sample --- clearing the entire intersection of the pruned $\varepsilon$-ball with the sampled cell $\kappa$ --- is larger than the probability of finding a \emph{bad} sample --- potentially leading to a cell split in traversing dimension. We then extend this result to outer cells of a pruned $\varepsilon$-ball.

We will first show, that the probability of finding a \emph{good} sample $P_\textnormal{good}(\kappa)$ is bounded from below independent of the actual cell $\kappa$ that is sampled and independent of the size of the pruned $\varepsilon$-ball. A \emph{good} sample clears the entire intersection of the pruned $\varepsilon$-ball with $\kappa$.

\blemma
Whenever an inner cell $\kappa$ of a pruned $\varepsilon$-ball $\overline{B}_\varepsilon$ with $\varepsilon<\hat{\varepsilon}$ is sampled ($\hat{\varepsilon}$ as seen in the result of Lemma~\ref{Lem:cleared_ball}), the probability of finding a \emph{good} sample $P_\textnormal{good}(\kappa)$ clearing the entire intersection of $\kappa$ and $\overline{B}_\varepsilon$ is bounded from below independent of $\kappa$ and independent of the actual choice of $\varepsilon$.
\elemma

\bproof
This follows directly from Lemmata~\ref{Lem:sample_bound} and \ref{Lem:cleared_ball}.
The lower bound is
$$
P_\textnormal{good}(\kappa)\ge \hat{\varepsilon}^n.
$$
For an inner cell of $\overline{B}_\varepsilon$ we see that in traversing dimension $i$ it holds $D^\kappa_i=D^I_i$ in terms of the proof of Lemmata~\ref{Lem:sample_bound} and \ref{Lem:sample_bound_tight} and consequentially the lower bound can be raised to
$$
P_\textnormal{good}(\kappa)\ge \hat{\varepsilon}^{n-1}.
$$
\eproof

\blemma
\label{Lem:bad_sample}
Whenever an inner cell $\kappa$ of a pruned $\varepsilon$-ball $\overline{B}_\varepsilon$ with $\varepsilon<\hat{\varepsilon}$ is sampled, the probability of finding a \emph{bad} sample $P_\textnormal{bad}(\kappa)$ potentially leading to a split of $\kappa$ in traversing dimension is bounded from above. This upper bound goes to zero with the cell width of $\kappa$ in traversing dimension for all but a finite number of cells.
\elemma

To prove Lemma \ref{Lem:bad_sample} we first have to introduce some more useful notation. We then bound the volume where such a bad sample may appear and show that the probability of sampling in this volume goes to zero for all but a finite number of inner cells $\kappa$.

\noindent For a set $A\in\mathbb{R}^n$ let $A_{|_i}$ denote its projection onto the $(n-1)$-dimensional subspace perpendicular to the $i$-th dimension, i.e.
$$A_{|_i}=\left\{x\in\mathbb{R}^{n-1}|\exists x'\in A, x = (x'_1,\dots,x'_{i-1},x'_{i+1},x'_n)\right\}.$$

\noindent Let $A_{|_{i,x_i}}$ or shorter $A_{|_{x_i}}$ denote its $(n-1)$-dimensional cross section through $x_i$ perpendicular to the $i$-th dimension, i.e.
$$A_{|_{x_i}}=\left\{x\in\mathbb{R}^{n-1}|\exists x'\in A, x'_i=x_i, x = (x'_1,\dots,x'_{i-1},x'_{i+1},x'_n)\right\}.$$

\noindent Let furthermore $A_{|_{i,x_i\pm\varepsilon}}$ or shorter $A_{|_{x_i\pm\varepsilon}}$ be defined by
\small
$$A_{|_{x_i\pm\varepsilon}}=\left\{x\in\mathbb{R}^{n-1}|\exists x'\in A, x'_i\in[x_i-\varepsilon, x_i+\varepsilon], x = (x'_1,\dots,x'_{i-1},x'_{i+1},x'_n)\right\}.$$

\begin{figure}
    \center
    \begin{tabular}{ccc}
    \subfloat[$A_{|_i}$]{\includegraphics[width=0.3\linewidth]{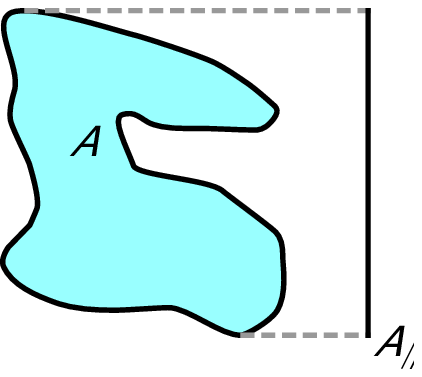}}&
    \subfloat[$A_{|_{x_i}}$]{\includegraphics[width=0.28\linewidth]{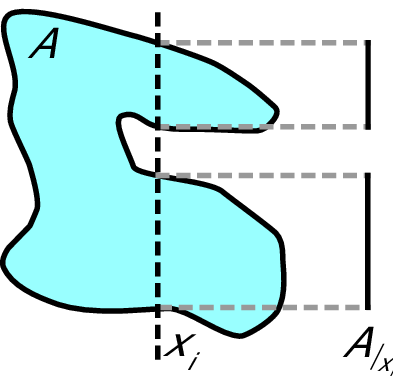}}&
    \subfloat[$A_{|_{x_i\pm\varepsilon}}$]{\includegraphics[width=0.3\linewidth]{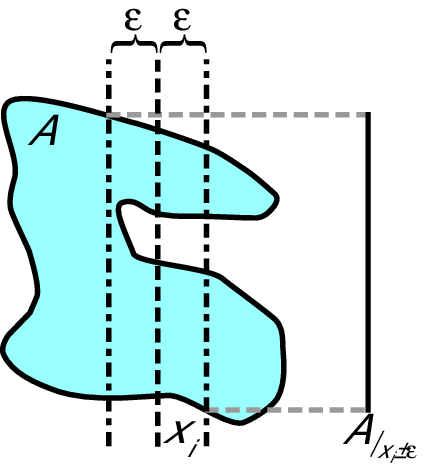}}
    \end{tabular}
    \caption[Projections]
            {(a): $A_{|_i}$ is the projection of $A$ (light blue set) onto the $(n-1)$-dimensional subspace perpendicular to the $i$-th dimension.
             (b): $A_{|_{x_i}}$ is the $(n-1)$-dimensional cross section of $A$ through $x_i$ (dashed black line) perpendicular to the $i$-th dimension.
             (c): $A_{|_{x_i\pm\varepsilon}}$ is the projection of the $2\varepsilon$-wide cross section of $A$ through $x_i$ perpendicular to the $i$-th dimension. }
    \label{Fig:proj}
\end{figure}

\normalsize
\noindent $A_{|_{x_i\pm\varepsilon}}$ could be described as being the $(n-1)$-dimensional projection of the $2\varepsilon$-wide cross section through $x_i$ perpendicular to the $i$-th dimension. Instead of $A_{|_{x_i\pm\varepsilon}}$ we sometimes use the notation $A_{|_{[x_i-\varepsilon, x_i+\varepsilon]}}$ later on. Please see Figure \ref{Fig:proj} for an illustration of the three projections.

\blemma
If $A$ is a compact semi-algebraic set, also $A_{|_i}$, $A_{|_{x_i}}$ and $A_{|_{x_i\pm\varepsilon}}$ are compact semi-algebraic sets.
\elemma

\bproof
The Tarski-Seidenberg theorem (see, for example, \cite{coste2000} for reference) states that semi-algebraic sets are closed under projection. Hence, $A_{|_i}$ is semi-algebraic. The canonical projection $\mathbb{R}^n\rightarrow\mathbb{R}^{n-1}$ is a continuous map and hence the image of any compact set is compact.

\noindent$A_{|_{x_i}}=(A\cap\left\{x'\in \mathbb{R}^n|f(x')=x'_i-x_i=0\right\})_{|_i}$ is the projection of the intersection of two semi-algebraic sets. Hence, it is semi-algebraic. The set $\left\{x'\in \mathbb{R}^n|f(x')=x'_i-x_i=0\right\}$ is closed in $\mathbb{R}^n$ and the intersection of a closed and a compact set is compact. Since the canonical projection of a compact set is compact, $A_{|_{x_i}}$ is compact in $\mathbb{R}^{n-1}$.

\noindent$A_{|_{x_i\pm\varepsilon}}=(A\cap\left\{x'\in \mathbb{R}^n|f(x')= x'_i-(x_i-\varepsilon)\ge0, g(x')=x'_i-(x_i+\varepsilon)\le0\right\})_{|_i}$ is the projection of  the intersection of two semi-algebraic sets. Hence, it is semi-algebraic. $\left\{x'\in \mathbb{R}^n|f(x')= x'_i-(x_i-\varepsilon)\ge0, g(x')=x'_i-(x_i+\varepsilon)\le0\right\}$ is closed in $\mathbb{R}^n$. Following the argument directly above, $A_{|_{x_i\pm\varepsilon}}$ is compact in $\mathbb{R}^{n-1}$.
\eproof

As a comment: $\mathcal{C}_\textnormal{obst}$ and its boundary $\delta\mathcal{C}_\textnormal{obst}$ are closed and we are merely interested in the conservation of the closedness property. Since there are examples where the canonical projection of a closed set is not closed, we have to require compact sets $A$. Since $\mathcal{C}_\textnormal{obst}\subset\mathcal{C}=[0,1]^n$, $\mathcal{C}_\textnormal{obst}$ and $\delta\mathcal{C}_\textnormal{obst}$ are not only closed but also bounded and hence compact.

With the right tools in place we can now take a look at the \emph{bad} samples that potentially lead to a cell split perpendicular to the traversing dimension of a cell. Please recall that we examine the position of collision-free samples found in the sampling step. For these we can assume a uniform distribution over the cell that is sampled. We do not assume to have any knowledge about colliding samples in the cell that is sampled. These can stem from both probabilistic sampling or from collision checking while checking a continuous path in the deterministic part of the algorithm \texttt{<a>}. Therefore, we make a worst case analysis and assume the colliding samples to be found in the worst possible places. Thus, if a collision-free sample is found that might potentially lead to a split in the traversing dimension if a colliding sample is found in the wrong place, we assume this colliding sample to be found exactly there.

If a collision-free sample $q_\textnormal{f}$ shall potentially lead to a cell split in traversing dimension $i$, there has to exist a colliding configuration $q_\textnormal{c}$ in the $i$-th split sector of $q_\textnormal{f}$, $S^i(q_\textnormal{f})$. But if a colliding configuration exists in the $i$-th split sector of $q_\textnormal{f}$ then also at least one configuration on the $\mathcal{C}_\textnormal{obst}$ boundary has to exist in $S^i(q_\textnormal{f})$. This leads to the following lemma:

\blemma
An inner cell $\kappa$ of a pruned $\varepsilon$-ball can be split in the traversing dimension only if a collision-free sample $q_\textnormal{f}$ is found with the $i$-th split sector of $q_\textnormal{f}$ containing a fraction of the boundary of $\mathcal{C}_\textnormal{obst}$: $S^i(q_\textnormal{f})\cap\delta\mathcal{C}_\textnormal{obst}\neq\emptyset$.
\elemma
\bproof
By definition of the split sectors, a cell is split in any dimension $i$ only if a colliding sample is found in the $i$-th split sector of a collision-free sample or if a collision-free sample is found in the $i$-th split sector of a colliding sample. Since $a$ is in the $i$-th split sector of $b$ is equivalent to $b$ is in the $i$-th split sector of $a$, it is sufficient to examine the $i$-th split sector of $q_\textnormal{f}$ only. For the cell to be potentially split in the traversing dimension, there has to be a colliding configuration $q_\textnormal{c}\in\kappa$ in the $i$-th split sector of $q_\textnormal{f}$. From the definition of the split sector we get for $q_\textnormal{c}$ in $S^i(q_\textnormal{f})$ also the straight line connection $L=q_\textnormal{f}+t(q_\textnormal{c}-q_\textnormal{f}), t\in[0,1]$ of these two samples is in $S^i(q_\textnormal{f})$. Hence also the point $\delta\mathcal{C}_\textnormal{obst}\cap L$ is in $S^i(q_\textnormal{f})$.
\eproof

 Thus, a sample is \emph{bad} only if its $i$-th split sector contains a fraction of the boundary of the $\mathcal{C}$-space obstacles $\delta\mathcal{C}_\textnormal{obst}$. But then from the symmetry of the split sector relation we get that the set of \emph{bad} samples is the collision-free subset of the union of all split sectors originating at configurations on $\delta\mathcal{C}_\textnormal{obst}$ in $\kappa$.

\blemma
The set of \emph{bad} configurations of an inner cell $\kappa$ of a pruned $\varepsilon$-ball is given by $Q=\kappa\cap\mathcal{C}_\textnormal{free}\cap\bigcup_{q\in\delta\mathcal{C}_\textnormal{obst}\cap\kappa}S^i(q)$.
\elemma
\bproof
This comes straightforwardly from the definition of a \emph{bad} sample. It has to (a): be a configuration of $\kappa$, (b): it has to be collision-free and (c): its $i$-th split sector has to contain a fraction of $\delta\mathcal{C}_\textnormal{obst}$ in $\kappa$ and by symmetry it has to be contained in the $i$-th split sector of a $\delta\mathcal{C}_\textnormal{obst}$ configuration in $\kappa$.
\eproof

\begin{figure}
    \center
    \begin{tabular}{ccc}
    \subfloat[$Q$]{\includegraphics[width=0.1426\linewidth]{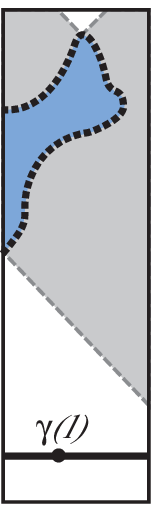}}&\hphantom{XXX}&
    \subfloat[Upper bound on $Q$]{\includegraphics[width=0.4\linewidth]{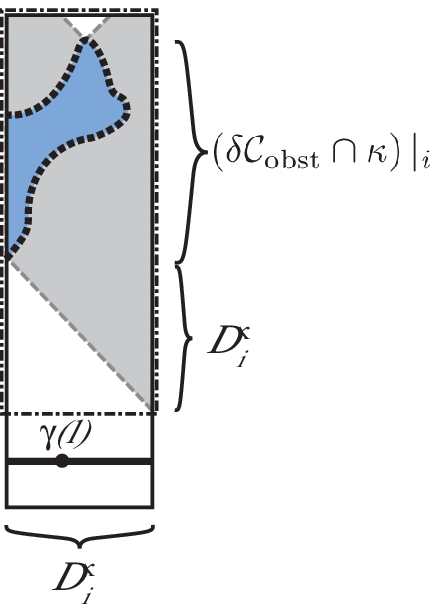}}
    \end{tabular}
    \caption[Upper bound on probability of finding a bad sample]
    {(a): shows a possibly occupied inner cell $\kappa$ of a pruned $\varepsilon$-ball. The path $\gamma$ (bold black line) traverses $\kappa$ horizontally. $\mathcal{C}_\textnormal{obst}$ (blue area); $\delta\mathcal{C}_\textnormal{obst}$ (dotted line); The set of \emph{bad} configurations $Q$ (grey area) is bounded by the boundary of $\kappa$, $\delta\mathcal{C}_\textnormal{obst}$ and the outer split sector boundaries (dashed grey lines). (b) depicts the derivation of an upper bound of this volume (dash-dotted rectangle). In traversing dimension $i$ it simply covers the whole cell. In the hyperplane perpendicular to the traversing dimension it covers the projection of $\delta\mathcal{C}_\textnormal{obst}$ in $\kappa$ onto it, widened by a $D^\kappa_i$-ball and restricted to the projection of $\kappa$. The probability of finding a \emph{bad} sample is smaller than the probability of finding a sample within this rectangle, so
    $P_\textnormal{bad}(\kappa)\le\frac{D^\kappa_i\times\mu^{n-1}(((\delta\mathcal{C}_\textnormal{obst}\cap\kappa)_{|_i}\oplus B^{n-1}_{D^\kappa_i})\cap\kappa_{|_i})}{\mu(\kappa)}$.}
    \label{Fig:bad_area}
\end{figure}

See Figure \ref{Fig:bad_area}.a for an example. The grey area corresponds to $Q$, the set of \emph{bad} configurations. But now we can bound the probability of finding such a sample from above. This derivation is depicted in \ref{Fig:bad_area}.b.

\blemma
\label{Lem:bad_bound_inner}
Let $\kappa$ be an inner cell of a pruned $\varepsilon$-ball $\overline{B}_\varepsilon$ with cell width $D^\kappa_i$ in traversing dimension $i$. The probability of finding a \emph{bad} sample $q_\textnormal{f}$ in $\kappa$ can be bounded from above by
\begin{align*}
 P_\textnormal{bad}(\kappa)&\le\frac{D^\kappa_i\times\mu^{n-1}(((\delta\mathcal{C}_\textnormal{obst}\cap\kappa)_{|_i}\oplus B^{n-1}_{D^\kappa_i})\cap\kappa_{|_i})}{\mu(\kappa)}\\
&=\frac{\mu^{n-1}(((\delta\mathcal{C}_\textnormal{obst}\cap\kappa)_{|_i}\oplus B^{n-1}_{D^\kappa_i})\cap\kappa_{|_i})}{\mu^{n-1}(\kappa_{|_i})},
\end{align*}
 i.e. the volume where a sample could lead to a split in the traversing dimension is not greater than the intersection of the boundary of $\mathcal{C}_\textnormal{obst}$ with $\kappa$ projected onto the hyperplane perpendicular to dimension $i$ and widened by an $n-1$-dimensional $D^\kappa_i$-ball times the cell width in traversing dimension $D^\kappa_i$.
\elemma

\bproof
Since $\kappa$ is an inner cell of the pruned $\varepsilon$-ball $\overline{B}_\varepsilon$ it has width $D^\kappa_i<2\varepsilon$ in the traversing dimension $i$. The $i$-th split sector of $q_\textnormal{f}$ is given by $S^i(q_\textnormal{f})=\left\{q|q\in\kappa, d_i(q-q_\textnormal{f})\ge d_j(q-q_\textnormal{f}), 1\le i\neq j\le n\right\}$. Thus, the distance between $q_\textnormal{f}$ and any colliding sample $q_\textnormal{c}$ in the split sector in any other dimension than $i$ must be not greater than the actual distance in dimension $i$ which cannot be greater than $D^\kappa_i$. But then $q_\textnormal{f}$ must have been found at a distance of $d_j\le d_i$ from a configuration on the boundary of $\mathcal{C}_\textnormal{obst}$ which implies $q_{\textnormal{f}|_i}\in(\delta\mathcal{C}_\textnormal{obst}\cap\kappa)_{|_i}\oplus B_{D^\kappa_i}^{n-1}$. From $q_\textnormal{f}\in\kappa$ it follows $q_{\textnormal{f}|_i}\in((\delta\mathcal{C}_\textnormal{obst}\cap\kappa)_{|_i}\oplus B_{D^\kappa_i}^{n-1})\cap\kappa_{|_i}$.
\eproof

\begin{figure}
    \center
    \subfloat[]{\includegraphics[width=0.6\linewidth]{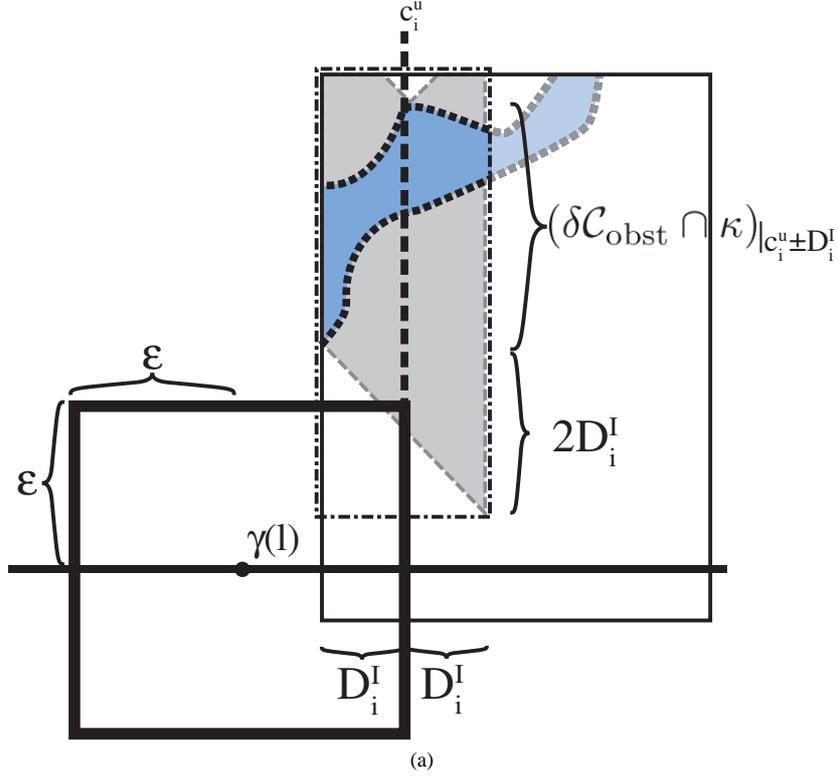}}
    \caption[Bound on probability of finding a \emph{good} or \emph{bad} sample]
    {The figure shows the derivation of an upper bound on the volume where a \emph{bad} sample $q_\textnormal{f}$ might be found (dash-dotted rectangle) potentially leading to a cell split in traversing dimension of an outer cell $\kappa$ (thin black rectangle) of a pruned $\varepsilon$-ball $\overline{B}_\varepsilon(\gamma(l))$ cutting through $\overline{B}_\varepsilon(\gamma(l))$. The corresponding non-pruned $\varepsilon$-ball $B_\varepsilon(\gamma(l))$ is given by the bold black rectangle. }
    \label{Fig:bad_outer}
\end{figure}

We now carry over the same argument to an outer cell of a pruned $\varepsilon$-ball $\overline{B}_\varepsilon$ where we get a very similar result. Here it is important that we want to bound the probability of finding a \emph{bad} sample that not only splits the cell $\kappa$ in traversing dimension $i$ but that this split also has to cut through $\overline{B}_\varepsilon$. Keep in mind that we want to bound the number of samples in $\overline{B}_\varepsilon$ by showing that it will not intersect with more and more cells as PCD progresses. If the split cuts through $\kappa$ in the traversing dimension but not through $\overline{B}_\varepsilon$ the number of cells that intersect with $\overline{B}_\varepsilon$ simply does not change. Please see Figure~\ref{Fig:bad_outer} for a sketch of the derivation of the upper bound for an outer cell.

\blemma
\label{Lem:bad_bound_outer}
Let $\kappa$ be an upper (lower) outer cell of a pruned $\varepsilon$-ball $\overline{B}_\varepsilon$ with the intersection $I=\kappa\cap\overline{B}_\varepsilon$ having width $D^I_i$ in traversing dimension $i$. The probability of finding a \emph{bad} sample $q_\textnormal{f}$ in $\kappa$ that potentially leads to a cell split in traversing dimension cutting through $\overline{B}_\varepsilon$ can be bounded from above by
 $$P_\textnormal{bad}(\kappa)\le\frac{\min(2D^I_i, D^\kappa_i)\times\mu^{n-1}(((\delta\mathcal{C}_\textnormal{obst}\cap\kappa)_{|_{c^\textnormal{u}_i\pm D^I_i}}\oplus B^{n-1}_{2D^I_i})\cap\kappa_{|_i})}{\mu(\kappa)},$$
  with $c^\textnormal{u}_i$ being the $i$-th coordinate of the upper defining vertex of $\overline{B}_\varepsilon$, $c^\textnormal{u}$ (replaced by $c^\textnormal{l}_i$ for a lower outer cell). Thus, in dimension $i$, the sample may be found in the intersection $I$ or at a maximum distance of the width of this intersection outside $\overline{B}_\varepsilon$.
\elemma
\bproof
If the potential split shall cut through $\overline{B}_\varepsilon$ in traversing dimension $i$, $q^\textnormal{f}$ has to be found on one side of the split and a colliding sample $q^\textnormal{c}$ on the other side at the same distance. This limits the area where $q^\textnormal{f}$ could be found to $[c_i^\textnormal{u} - D^I_i,  c_i^\textnormal{u} + D^I_i]$ in dimension $i$. $q^\textnormal{f}_i \ge c^\textnormal{u}_i-D^I_i$ is obvious from $q^\textnormal{f}\in \kappa$. For the upper bound we get: For any colliding sample $q^\textnormal{c}$ in $\kappa$ it obviously holds $q^\textnormal{c}_i \ge c^\textnormal{u}_i-D^I_i$. But then, if $q^\textnormal{f}_i > c^\textnormal{u}_i+D^I_i$ for the split location it holds $x^\textnormal{split}_i = (q^\textnormal{c}_i+q^\textnormal{f}_i)/2 > c^\textnormal{u}_i$ which is outside the intersection $I$. Thus, $q^\textnormal{f}_i \le c^\textnormal{u}_i+D^I_i$ and consequentially, the width of the volume where a bad sample might be found is less than $2D^\textnormal{I}_i$. If the width of $\kappa$ in dimension $i$ is smaller than $2D^I_i$, the width of the volume where a \emph{bad} sample might be found is $D^\kappa_i$.

Following the same argument as in the proof of Lemma \ref{Lem:bad_bound_inner}, in the other dimensions the volume can be bounded by $((\delta\mathcal{C}_\textnormal{obst}\cap\kappa)_{|_{c^\textnormal{u}_i\pm D^I_i}}\oplus B^{n-1}_{2D^I_i})\cap\kappa_{|_i})$. Please observe that the projection of $\delta\mathcal{C}_\textnormal{obst}$ in the relevant area is widened by a $2D^I_i$-ball. Accordingly for a lower outer cell by replacing $c^\textnormal{u}$ with $c^\textnormal{l}$.
\eproof

\blemma
Let $\tilde\varepsilon = 4\varepsilon$. Then for any inner or outer cell $\kappa$ of a pruned $\varepsilon$-ball $\overline{B}_\varepsilon$ it holds
$$P_\textnormal{bad}(\kappa)\le\frac{\mu^{n-1}(\delta\mathcal{C}_\textnormal{obst}|_{x_i \pm \tilde\varepsilon}\oplus B^{n-1}_{\tilde\varepsilon})}{\mu^{n-1}(\kappa_{|_i})}$$
where for an inner cell $x_i = (c^{\kappa\textnormal{l}}_i+c^{\kappa\textnormal{u}}_i)/2$ is the cell center in dimension $i$ or for a lower or upper outer cell $x_i = c^{\overline{B}_\varepsilon\textnormal{l}}_i$ or $x_i = c^{\overline{B}_\varepsilon\textnormal{u}}_i$ is the lower or upper boundary of $\overline{B}_\varepsilon$ in dimension $i$, respectively.
\elemma

\bproof
$D^\kappa_i\le2\varepsilon<4\varepsilon$ from the definition of an inner cell of the pruned $\varepsilon$-ball with diameter $2\varepsilon$. For the outer cell, $D^I_i\le 2\varepsilon$ and hence $2D^I_i\le 4\varepsilon$.  For the inner cell, we have to show that
$$
\mu^{n-1}(((\delta\mathcal{C}_\textnormal{obst}\cap\kappa)_{|_i}\oplus B^{n-1}_{D^\kappa_i})\cap\kappa_{|_i}) \le \mu^{n-1}(\delta\mathcal{C}_\textnormal{obst}|_{x_i \pm \tilde\varepsilon}\oplus B^{n-1}_{\tilde\varepsilon})
$$
Taking a look at the volumes that are measured on both sides of the equation we can see that
\begin{align*}
((\delta\mathcal{C}_\textnormal{obst}\cap\kappa)_{|_i}\oplus B^{n-1}_{D^\kappa_i})\cap\kappa_{|_i} &\subseteq (\delta\mathcal{C}_\textnormal{obst}\cap\kappa)_{|_i}\oplus B^{n-1}_{D^\kappa_i}\\
&\subseteq (\delta\mathcal{C}_\textnormal{obst}\cap\kappa)_{|_i}\oplus B^{n-1}_{\tilde\varepsilon}\\
&\subseteq \delta\mathcal{C}_\textnormal{obst}|_{x_i \pm \tilde\varepsilon}\oplus B^{n-1}_{\tilde\varepsilon}.
\end{align*}
The first inclusion comes from deleting the last intersection which can make the set only bigger. The second inclusion uses $\tilde\varepsilon=4\varepsilon\ge D^\kappa_i$ and $A\oplus B \subseteq A \oplus B'$ if $B\subseteq B'$. For the third inclusion we recall that $x_i = (c^{\kappa\textnormal{l}}_i+c^{\kappa\textnormal{u}}_i)/2$ is the center of $\kappa$ in dimension $i$. Now $x_i - \tilde\varepsilon < c^{\kappa\textnormal{l}}_i$ and $x_i + \tilde\varepsilon > c^{\kappa\textnormal{u}}_i$ and thus we get from the definition of $|_{x_i\pm \tilde{\varepsilon}}$ that $(\delta\mathcal{C}_\textnormal{obst}\cap\kappa)_{|_i}\subseteq\delta\mathcal{C}_\textnormal{obst}|_{x_i \pm \tilde\varepsilon}$. And from the monotonicity of the measure $\mu^{n-1}$ we get
$$
\mu^{n-1}(((\delta\mathcal{C}_\textnormal{obst}\cap\kappa)_{|_i}\oplus B^{n-1}_{D^\kappa_i})\cap\kappa_{|_i}) \le \mu^{n-1}(\delta\mathcal{C}_\textnormal{obst}|_{x_i \pm \tilde\varepsilon}\oplus B^{n-1}_{\tilde\varepsilon})
$$

\noindent For the upper outer cell, we have to show that
$$
\mu^{n-1}(((\delta\mathcal{C}_\textnormal{obst}\cap\kappa)_{|_{c^\textnormal{u}_i\pm D^I_i}}\oplus B^{n-1}_{2D^I_i})\cap\kappa_{|_i}) \le \mu^{n-1}(\delta\mathcal{C}_\textnormal{obst}|_{x_i \pm \tilde\varepsilon}\oplus B^{n-1}_{\tilde\varepsilon})
$$
Again taking a look at the volumes that are measured on both sides of the equation we can see that
\begin{align*}
((\delta\mathcal{C}_\textnormal{obst}\cap\kappa)_{|_{c^\textnormal{u}_i\pm D^I_i}}\oplus B^{n-1}_{2D^I_i})\cap\kappa_{|_i} & \subseteq (\delta\mathcal{C}_\textnormal{obst}\cap\kappa)_{|_{c^\textnormal{u}_i\pm D^I_i}}\oplus B^{n-1}_{2D^I_i}\\
& \subseteq \delta\mathcal{C}_\textnormal{obst}|_{c^\textnormal{u}_i\pm D^I_i}\oplus B^{n-1}_{2D^I_i}\\
& \subseteq \delta\mathcal{C}_\textnormal{obst}|_{c^\textnormal{u}_i\pm D^I_i}\oplus B^{n-1}_{\tilde\varepsilon}\\
& \subseteq \delta\mathcal{C}_\textnormal{obst}|_{x_i \pm \tilde\varepsilon}\oplus B^{n-1}_{\tilde\varepsilon}.
\end{align*}
The inclusions work analog to those for the inner cell.
\eproof

\begin{figure}
    \center
    \begin{tabular}{ccccc}
    \subfloat[]{\includegraphics[width=0.1426\linewidth]{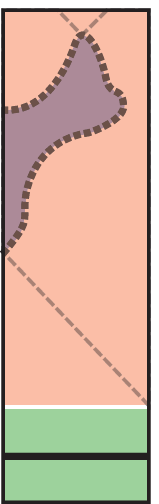}}&
    \subfloat[]{\includegraphics[width=0.1426\linewidth]{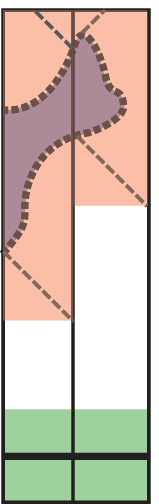}}&
    \subfloat[]{\includegraphics[width=0.1426\linewidth]{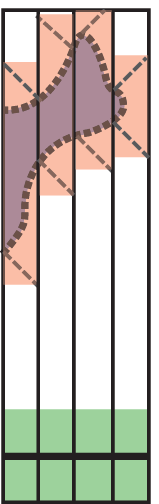}}&
    \subfloat[]{\includegraphics[width=0.1426\linewidth]{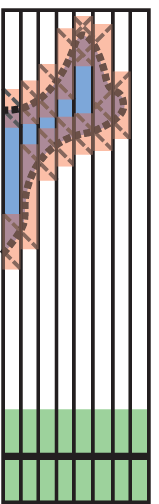}}&
    \subfloat[]{\includegraphics[width=0.1426\linewidth]{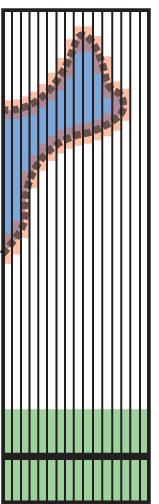}}
    \end{tabular}
    \caption[Bound on probability of finding a \emph{good} or \emph{bad} sample]
    {The figures show the upper bound on the volume where a \emph{bad} sample might be found (shaded red) potentially leading to a cell split in traversing dimension and the lower bound on the volume where a \emph{good} sample might be found (shaded green) clearing the entire intersection of $\kappa$ and the pruned $\varepsilon$-ball $\overline{B}_\varepsilon$. From figures (a) to (e) the cells are split in half in traversing dimension. While the lower bound on the probability of finding a \emph{good} sample stays constant, the probability of finding a \emph{bad} sample decreases to zero with the cell width in traversing dimension for almost all cells. Path $\gamma$: bold black line; obstacles: shaded blue, bounded by dotted line.}
    \label{Fig:bounds_development}
\end{figure}

Now we show that this upper bound on the probability of finding a \emph{bad} sample goes to zero with $\varepsilon$ for almost all $x_i$. Figure~\ref{Fig:bounds_development} illustrates the contrast of constant lower bound on finding a \emph{good} sample and diminishing upper bound on finding a \emph{bad} sample when the size of the pruned $\varepsilon$-balls of the finite covering of $\gamma$ goes to zero. The figure shows how the upper bound evolves when cells are split in equal halves just for illustrative purposes. This exact halving has no basis in the algorithm of PCD. With respect to the lower bound on finding a \emph{good} sample the illustration is not to scale in terms of Lemma~\ref{Lem:cleared_ball}.

Unfortunately, the upper bound on the probability of finding a \emph{bad} sample does not go to zero for all $x_i$. For some $x_i$ the probability has some positive lower limit. The problems arise where $\delta\mathcal{C}_\textnormal{obst}$ runs exactly perpendicular to the traversing dimension $i$ over an area with nonzero $\mu^{n-1}$. The third cell from the left in Figure~\ref{Fig:bad_area}.e gives a rough idea about this issue.

In the following, we use some results from basic algebra: Let $H_{{x_i},i}$ or shorter $H_{x_i}$ be a hyperplane perpendicular to dimension $i$ through $x_i$. If a single polynomial $f: \mathbb{R}^n\rightarrow\mathbb{R}$ is zero on a subset of $H_{x_i}$ with nonzero measure $\mu^{n-1}$ it is zero on the entire hyperplane. Thus $f$ reduces to a polynomial in $x_i$ only. A polynomial in one variable which is not the trivial zero polynomial has at most $d$ real zeros, where $d$ is the degree of the polynomial. Thus $f$ can be zero on at most $d$ hyperplanes perpendicular to dimension $i$. Since $\delta\mathcal{C}_\textnormal{obst}$ is a semi-algebraic set defined by a finite number of polynomials with finite degree, the number of hyperplanes $H_{x_i}$ with $\mu^{n-1}((H_{x_i}\cap\delta\mathcal{C}_\textnormal{obst})_{|_i})>0$ is finite. According to the definition given above for $(H_{x_i}\cap\delta\mathcal{C}_\textnormal{obst})_{|_i}$ we can write ${\delta\mathcal{C}_\textnormal{obst}}_{|_{x_i}}$.

In our analysis of the upper bound on finding a \emph{bad} sample we therefore restrict ourselves -- for the moment -- to cells $\kappa$ that do not contain areas on $\delta\mathcal{C}_\textnormal{obst}$ exactly perpendicular to the traversing dimension $i$ with $\mu^{n-1} > 0$. With $X_i=\{x_i|\mu^{n-1}({\delta\mathcal{C}_\textnormal{obst}}_{|_{x_i}})>0\}$ being the finite set of $x_i$ where $\delta\mathcal{C}_\textnormal{obst}$ runs perpendicular to dimension $i$ on an area with nonzero $\mu^{n-1}$ we define
$$\Delta^i\mathcal{C}_\textnormal{obst} =\textnormal{cl}(\delta\mathcal{C}_\textnormal{obst}\setminus \bigcup_{x_i\in X_i} H_{x_i}).$$
Thus with an uppercase $\Delta^i$ we denote the boundary of $\mathcal{C}_\textnormal{obst}$ stripped by all points on hyperplanes $H_{x_i}$ with $\mu^{n-1}({\delta\mathcal{C}_\textnormal{obst}}\cap{H_{x_i}})>0$ adding back the limit points of this set to maintain a closed set.
Thereby, the following analysis is valid only for cells that do not intersect with such a Hyperplane $H_{x_i}$.

We now show that when we let the size of the pruned $\varepsilon$-balls around $\gamma$ go to zero, for all cells of a pruned $\varepsilon$-ball not intersecting with a hyperplane as defined above the upper bound on the probability of finding a \emph{bad} sample goes uniformly to zero, i.e. $\forall \nu >0,\ \exists \varepsilon > 0,\ \forall x_i\in[0, 1]: \mu^{n-1}(\Delta^i\mathcal{C}_\textnormal{obst}|_{x_i\pm\varepsilon}\oplus B^{n-1}_\varepsilon)<\nu$.

\blemma
\label{Lem:uniform_convergence}
For all $1 \le i \le n$,
$$f(x_i)=\mu^{n-1}(\Delta^i\mathcal{C}_\textnormal{obst}|_{x_i\pm\varepsilon}\oplus B^{n-1}_\varepsilon)\xrightarrow[\varepsilon\rightarrow0]{\textnormal{uniform}}0\triangleq F$$
on $[0,1]$.
\elemma
To prove Lemma~\ref{Lem:uniform_convergence}, we need the following intermediate results:

\blemma
\label{Lem:uniform_convergence_if}
If
\begin{align*}
\mu^{n-1}(\Delta^i\mathcal{C}_\textnormal{obst}|_{x_i\pm\tilde{\varepsilon}}\oplus B^{n-1}_\varepsilon)&\xrightarrow[\varepsilon\rightarrow0]{\textnormal{uniform}}
\mu^{n-1}(\Delta^i\mathcal{C}_\textnormal{obst}|_{x_i\pm\tilde{\varepsilon}})\ \textnormal{ and}\\
\mu^{n-1}(\Delta^i\mathcal{C}_\textnormal{obst}|_{x_i\pm\varepsilon})&\xrightarrow[\varepsilon\rightarrow0]{\textnormal{uniform}}0\ \textnormal{ then}\\
\mu^{n-1}(\Delta^i\mathcal{C}_\textnormal{obst}|_{x_i\pm\varepsilon}\oplus B^{n-1}_\varepsilon)&\xrightarrow[\varepsilon\rightarrow0]{\textnormal{uniform}}0.
\end{align*}
If the expression converges uniformly when letting one $\varepsilon$ go to zero and holding the other one fixed, then the expression converges uniformly also when letting both $\varepsilon$ go to zero.
\elemma

\bproof
From $f^1(x_i)\triangleq\mu^{n-1}(\Delta^i\mathcal{C}_\textnormal{obst}|_{x_i\pm\tilde{\varepsilon}}\oplus B^{n-1}_\varepsilon)\xrightarrow[\varepsilon\rightarrow0]{\textnormal{uniform}}
\mu^{n-1}(\Delta^i\mathcal{C}_\textnormal{obst}|_{x_i\pm\tilde{\varepsilon}})$ it follows that $\forall\eta>0,\ \exists\hat{\varepsilon}_1>0,\ \forall\hat{\varepsilon}_1>\varepsilon>0,\ \forall x_i\in[0,1]:$
$$
 \mu^{n-1}(\Delta^i\mathcal{C}_\textnormal{obst}|_{x_i\pm\tilde{\varepsilon}}\oplus B^{n-1}_\varepsilon)-\mu^{n-1}(\Delta^i\mathcal{C}_\textnormal{obst}|_{x_i\pm\tilde{\varepsilon}})<\eta.
$$

From $f^2(x_i)\triangleq\mu^{n-1}(\Delta^i\mathcal{C}_\textnormal{obst}|_{x_i\pm\varepsilon})\xrightarrow[\varepsilon\rightarrow0]{\textnormal{uniform}}0$ it follows that $\forall\eta>0,\ \exists\hat{\varepsilon}_2>0,\  \forall\hat{\varepsilon}_2>\varepsilon>0,\ \forall x_i\in[0,1]:$
\begin{align*}
 \mu^{n-1}(\Delta^i\mathcal{C}_\textnormal{obst}|_{x_i\pm\varepsilon})&<\eta.
\end{align*}
But then it follows that $\forall\hat{\eta}=2\eta,\ \exists\hat{\varepsilon} = \min(\hat{\varepsilon}_1,\hat{\varepsilon}_2),\ \forall \varepsilon < \hat{\varepsilon},\ \forall x_i\in[0,1]:$
$$\mu^{n-1}(\Delta^i\mathcal{C}_\textnormal{obst}|_{x_i\pm\varepsilon}\oplus B^{n-1}_\varepsilon)<2\eta=\hat{\eta}$$
and therefore  $\mu^{n-1}(\Delta^i\mathcal{C}_\textnormal{obst}|_{x_i\pm\varepsilon}\oplus B^{n-1}_\varepsilon)\xrightarrow[\varepsilon\rightarrow0]{\textnormal{uniform}}0$.
\eproof

So to get uniform convergence for $f(x_i)=\mu^{n-1}(\Delta^i\mathcal{C}_\textnormal{obst}|_{x_i\pm\varepsilon}\oplus B^{n-1}_\varepsilon)$ on $[0,1]$ we have to proof uniform convergence for $f^1$ and $f^2$. We switch to the standard form for proving convergence of a sequence of functions and replace $\varepsilon$ by $1/p$ and let $p$ go to infinity instead of $\varepsilon$ to zero: $f^1_p(x_i) = f^1(1/p,x_i) = f^1(\varepsilon,x_i), f^2_p(x_i) = f^2(1/p,x_i) = f^2(\varepsilon,x_i)$.

\blemma
\label{Lem:f2_converges_uniformly}
Let $f^2_p(x_i) = \mu^{n-1}(\Delta^i\mathcal{C}_\textnormal{obst}|_{x_i\pm 1/p})$, then
$${f^2_p(x_i)\xrightarrow[p\rightarrow\infty]{\textnormal{uniform}}0\triangleq F^2(x_i)}\ \textnormal{ on  }\ [0,1].$$
\elemma
\bproof
Dini's theorem (see for example \cite{bartle00} for reference) states that if a series of continuous functions ${(f^2_p:X\rightarrow\mathbb{R})_{p\in\mathbb{N}}}$ converges pointwise monotonously to a continuous $F^2$ on a compact set $X$, then $f^2_p$ converges also uniformly to $F^2$ on $X$.

We investigate $f^2_p$ on $x_i\in[0,1]$ which is closed and bounded and hence compact. For any $x_i\in[0,1]$ let $a^{x_i}_p=\Delta^i\mathcal{C}_\textnormal{obst}|_{x_i\pm 1/p}\in\mathcal{C}_{|_i}=[0,1]^{n-1}$. From the definition of the $\varepsilon$-wide projection we have $\Delta^i\mathcal{C}_\textnormal{obst}|_{x_i\pm 1/p} \supseteq \Delta^i\mathcal{C}_\textnormal{obst}|_{x_i\pm 1/(p+1)}$. Thus $\left\{a_p\right\}_{p\in\mathbb{N}}$ is a decreasing sequence of sets. For this decreasing sequence of subsets of $\mathcal{C}_{|_i}$ the Lebesgue measure of the limit is the limit of the Lebesgue measures and hence $\forall x_i\in[0,1],\ \lim_{p\rightarrow\infty}\mu^{n-1}(a_p^{x_i})= 0$. Thus $F^2=0$ and $f^2_p\xrightarrow[p\rightarrow\infty]{\textnormal{pointwise}}F^2$.

What is left to show is that each $f^2_p$ is continuous. A function $f^2_p$ is continuous on $[0, 1]$ if for every $x^\textnormal{cont}_i\in[0,1]$ the limit $\lim_{x_i\rightarrow x_i^\textnormal{cont}} f^2_p(x_i)$ exists and is equal to $f^2_p(x_i^\textnormal{cont})$. This holds true if the limit of $\lim_{x_i\rightarrow x_i^\textnormal{cont}} f^2_p(x_i)-f^2_p(x^\textnormal{cont}_i)$ exists and is equal to zero.
Let $x_i=x_i^\textnormal{cont}+\varepsilon$. But then
\begin{align*}
f^2_p(x_i) &= \mu^{n-1}(\Delta^i\mathcal{C}_\textnormal{obst}|_{x_i\pm 1/p})\\
&= \mu^{n-1}(\Delta^i\mathcal{C}_\textnormal{obst}|_{[x_i-1/p,\  x_i+1/p]})\\
&= \mu^{n-1}(\Delta^i\mathcal{C}_\textnormal{obst}|_{[x^\textnormal{cont}_i+\varepsilon-1/p,\  x^\textnormal{cont}_i+\varepsilon+1/p]})\\
f^2_p(x^\textnormal{cont}_i) &= \mu^{n-1}(\Delta^i\mathcal{C}_\textnormal{obst}|_{x^\textnormal{cont}_i\pm 1/p})\\
&= \mu^{n-1}(\Delta^i\mathcal{C}_\textnormal{obst}|_{[x^\textnormal{cont}_i-1/p,\  x^\textnormal{cont}_i+1/p]}).
\end{align*}
Let now
\begin{align*}
a&\triangleq \mu^{n-1}(\Delta^i\mathcal{C}_\textnormal{obst}|_{[x^\textnormal{cont}_i+1/p,\  x^\textnormal{cont}_i+1/p+\varepsilon]})\ \ \textnormal{and}\\
b&\triangleq \mu^{n-1}(\Delta^i\mathcal{C}_\textnormal{obst}|_{[x^\textnormal{cont}_i-1/p,\  x^\textnormal{cont}_i-1/p+\varepsilon]}).
\end{align*}
Then it holds
\begin{align*}
f^2_p(x_i) &\le f^2_p(x^\textnormal{cont}_i)+a\\
f^2_p(x^\textnormal{cont}_i) &\le f^2_p(x_i)+b\\
f^2_p(x^\textnormal{cont}_i)-b &\le f^2_p(x_i)\le f^2_p(x^\textnormal{cont}_i)+a\\
-b&\le f^2_p(x_i) -f^2_p(x^\textnormal{cont}_i)\le a\\
|f^2_p(x_i) -f^2_p(x^\textnormal{cont}_i)|&\le\max(a,b).
\end{align*}
By showing that both $a$ and $b$ go to zero for $\varepsilon\rightarrow0$ we conclude that the limit exists and is equal to zero. Thus, the $f^2_p$ are continuous on $[0,1]$ for all $p$. But for $\varepsilon$ going to zero both $\Delta^i\mathcal{C}_\textnormal{obst}|_{[x^\textnormal{cont}_i+1/p,\  x^\textnormal{cont}_i+1/p+\varepsilon]}$ and $\Delta^i\mathcal{C}_\textnormal{obst}|_{[x^\textnormal{cont}_i-1/p,\  x^\textnormal{cont}_i-1/p+\varepsilon]}$ are decreasing sequences of subsets of $\mathcal{C}_{|_i}$ with the limit of the measure being the measure of the limit which are $\mu^{n-1}(\Delta^i\mathcal{C}_\textnormal{obst}|_{x^\textnormal{cont}_i-1/p})$ and $\mu^{n-1}(\Delta^i\mathcal{C}_\textnormal{obst}|_{x^\textnormal{cont}_i+1/p})$, respectively, which are both equal to zero.

Thus we have a sequence of continuous $f^2_p$ that converges pointwise monotonously to a continuous $F^2$ on a compact set $[0,1]$, and hence $f^2_p$ converges also uniformly to $F^2$ on $[0,1]$.
\eproof

\blemma
\label{Lem:f1_converges_uniformly}
Let $f^1_p(x_i) = \mu^{n-1}(\Delta^i\mathcal{C}_\textnormal{obst}|_{x_i\pm\tilde{\varepsilon}}\oplus B^{n-1}_{1/p})$, then $$f^1_p(x_i)\xrightarrow[p\rightarrow\infty]{\textnormal{uniform}}\mu^{n-1}(\Delta^i\mathcal{C}_\textnormal{obst}|_{x_i\pm\tilde{\varepsilon}})\triangleq F^1(x_i)$$
\elemma

\bproof
$\Delta^i\mathcal{C}_\textnormal{obst}|_{x_i\pm\tilde{\varepsilon}}$ is a closed semi-algebraic set. A closed set is equal to its closure. The closure of any set $A$ is given by $\textnormal{cl}(A)=\bigcap_{p\in\mathbb{N}}(A+B_{1/p})$ and hence for closed sets $A=\lim_{p\rightarrow\infty}A\oplus B_{1/p}$. For a decreasing sequence of subsets of $\mathcal{C}_{|_i}$ the Lebesgue measure of the limit is the limit of the Lebesgue measures and hence $\forall x_i \in [0,1], f^1_p(x_i)\xrightarrow[p\rightarrow\infty]{\textnormal{pointwise}}F^1(x_i)$. Since the measured set in $f_{p+1}$ is a subset of the measured set in $f^1_p$ for any $p$, the $f^1_p$ decrease monotonously.

\begin{figure}
    \center
    \subfloat[]{\includegraphics[width=0.5\linewidth]{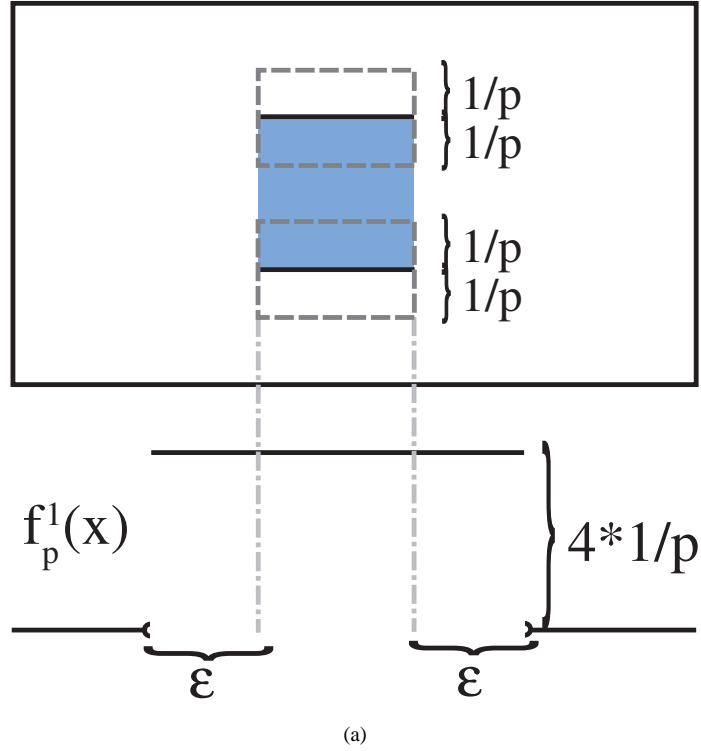}}
    \caption[$f^1_p$ not continuous]
    {The figure shows that the $f^1_p(x_i) = \mu^{n-1}(\Delta^i\mathcal{C}_\textnormal{obst}|_{x_i\pm\tilde{\varepsilon}}\oplus B^{n-1}_{1/p})$ are in general not continuous. The blue square in the center represents an obstacle in $\mathcal{C}$. $\delta\mathcal{C}_\textnormal{obst}$ is hence a rectangular frame around the blue area. $\Delta^i\mathcal{C}_\textnormal{obst}$ shown by black lines is this boundary stripped by all subsets that share an area with hyperplanes perpendicular to dimension $i$ with nonzero $\mu^{n-1}$.
    In the lower part of the figure, the graph of $f^1_p(x_i)$ is shown. Where $\Delta^i\mathcal{C}_\textnormal{obst}|_{x_i\pm\varepsilon}$ starts to contain a part of $\Delta^i\mathcal{C}_\textnormal{obst}$, $f^1_p$ is discontinuous.}
    \label{Fig:noncont}
\end{figure}

Following the argument in the proof of Lemma \ref{Lem:f2_converges_uniformly} the limit function $F^1$ is continuous. Since -- again  -- $[0,1]$ is compact we would need to show that each $f_p^1$ is continuous to use Dini's theorem to prove uniform convergence. Unfortunately, the $f_p^1$ are in general not continuous. See Figure \ref{Fig:noncont} for an example. However, the $f_p^1$ are discontinuous at finitely many $x^\textnormal{d}_k$, $1\le k\le k^\textnormal{d}$ only where the $x^\textnormal{d}_k$ are independent of $p$ (see Lemma~\ref{Lem:mu_finite_disc}). Furthermore, if $f_p^1$ is discontinuous in $x^\textnormal{d}_k$ from the left, then $f_p^1(x^\textnormal{d}_k)>f_p^1(x^\textnormal{d}_k-\varepsilon)$ for sufficiently small $\varepsilon$ and $f_p^1(x^\textnormal{d}_k)>f_p^1(x^\textnormal{d}_k+\varepsilon)$ for $f_p^1$ discontinuous in $x^\textnormal{d}_k$ from the right (see Lemma~\ref{Lem:disc_le_eps}).

Now partition $[0,1]$ into $k^\textnormal{d}+1$ compact intervals $I_k = [x^\textnormal{d}_k, x^\textnormal{d}_{k+1}]$ with $x^\textnormal{d}_0=0$ and $x^\textnormal{d}_{k^\textnormal{d}+1}=1$. We will prove uniform convergence for all $\hat{f}^1_{p,k}:I_k\rightarrow\mathbb{R}, \hat{f}^1_{p,k}(x_i)=f^1_p(x_i)$ for $x_i\in(x^\textnormal{d}_k,x^\textnormal{d}_{k+1})$ and $\hat{f}^1_{p,k}(x^\textnormal{d}_k)=\lim_{x_i'\searrow {x^\textnormal{d}_k}} f^1_p({x'}_i)$ and $\hat{f}^1_{p,k}(x^\textnormal{d}_{k+1})=\lim_{x_i'\nearrow {x^\textnormal{d}_{k+1}}} f^1_p({x_i}')$ in Lemma~\ref{Lem:f1_on_interval}.

But now for each $k$ the $\hat{f}^1_{p,k}$ are a sequence of continuous functions on a compact interval decreasing monotonously and converging pointwise to a continuous function $\hat{f}^1_k$ (see Lemma~\ref{Lem:f1_on_interval}). According to Dini's theorem, all $\hat{f}^1_{p,k}$ converge uniformly to $\hat{f}^1_k$. For all $x_i\in \textnormal{int} (I_k)$, $\hat{f}^1_{p,k}(x_i)=f_p^1(x_i)$. Thus we have for each $0\le k\le k_\textnormal{d}$ that $\forall \eta > 0, \ \exists P_k, \ \forall p > P_k, \ \forall x \in \textnormal{int} (I_k),\ |f^1_{p}(x)-f^1(x)|<\eta$. Similarly we get from the pointwise convergence of $f^1_p(x^\textnormal{d}_k)$ for each $0\le k\le k_{\textnormal{d}}+1$ that $\forall \eta > 0,\ \exists P'_{k},\ \forall p>P'_{k},\ |f^1_{p}(x^\textnormal{d}_{k})-f^1(x^\textnormal{d}_{k})|<\eta$. But then we have $\forall \eta>0, \ \exists P=\max(P_0,\dots,P_{k_\textnormal{d}},P'_0,\dots,P'_{k_\textnormal{d}+1}),\ \forall p > P, \ \forall x\in[0,1] \ \ |f^1_{p}(x)-f^1(x)|<\eta$ and hence uniform convergence of $f^1_p$ to $F^1$ on $[0,1]$.
\eproof

For proving the required continuity of $f^1_p$ we need a continuity concept for set-valued maps. In general, the continuity concept of single-valued functions does not trivially extend to higher dimensions. Instead one can define semi-continuity properties like upper and lower or inner and outer semi-continuity and a map is said to be continuous if both semi-continuity properties are fulfilled. See for example \cite{aubin09} and \cite{rockafellar98} for reference. Fortunately, for the compact-valued set-valued maps we want to examine, the continuity concepts are equivalent. In the following we use the Hausdorff continuity as it expresses continuity in the $\delta-\varepsilon$-notation similar to the continuity of single-valued functions.

 The Hausdorff metric gives a notion of distance between two sets. Using this metric, we can define continuity of set-valued maps with the standard $\delta-\varepsilon$-notation. We use $\eta$ instead of $\delta$ to avoid confusion with regard to the boundary of $\mathcal{C}$-space obstacles $\delta\mathcal{C}_\textnormal{obst}$.

\bdefinition[Hausdorff distance] By
$$d_\textnormal{H}:X\times X \rightarrow \mathbb{R}, d_\textnormal{H}(A,A')= \inf \left\{ \varepsilon>0 | A \subset A' + \widehat{B}_\varepsilon \wedge A' \subset A+\widehat{B}_\varepsilon\right\}$$ where $\widehat{B}_\varepsilon$ is the \emph{closed} $\varepsilon$-ball around zero we denote the Hausdorff distance.
\edefinition

The Hausdorff distance quantifies the largest distance from any element of $A$ to the respective closest element of $A'$ and vice versa.

\bdefinition[Hausdorff continuity]
A set-valued map $M: [0,1]\rightrightarrows \mathbb{R}^{n-1}$ is called Hausdorff continuous from the left in $x\in [0,1]$ if $\forall \eta >0 \ \ \exists \hat{\varepsilon}>0 \ \ \forall \hat{\varepsilon} >\varepsilon>0 \ \ d_\textnormal{H}(M(x-\varepsilon),M(x))<\eta$. Accordingly, with $d_\textnormal{H}(M(x+\varepsilon),M(x))<\eta$ for continuity from the right. If a set-valued map $M: [0,1]\rightrightarrows \mathbb{R}^{n-1}$ is Hausdorff continuous from the left and from the right in $x\in[0,1]$ it is said to be Hausdorff continuous in $x$. If such a map is not Hausdorff continuous from the left and / or from the right in $x$ it is said to be Hausdorff discontinuous in $x$.
\edefinition

\blemma
\label{Lem:ball_cont_mu_cont}
If the semi-algebraic set-valued map $M : [0,1]\rightrightarrows \mathbb{R}^{n-1}, M(x_i)=\Delta^i\mathcal{C}_\textnormal{obst}|_{x_i\pm\tilde{\varepsilon}}\oplus B^{n-1}_{1/p}$ is continuous in $x_i$ from the left (right), then the function $f^1_p(x_i)=\mu^{n-1}(\Delta^i\mathcal{C}_\textnormal{obst}|_{x_i\pm\tilde{\varepsilon}}\oplus B^{n-1}_{1/p})$ is continuous in $x_i$ from the left (right).
\elemma

\bproof
Assume $f^1_p(x_i)=\mu^{n-1}(\Delta^i\mathcal{C}_\textnormal{obst}|_{x_i\pm\tilde{\varepsilon}}\oplus B^{n-1}_{1/p})$ is discontinuous from the left in $x_i$. Then $\exists\eta\ \ \forall\hat\varepsilon>0\ \ \exists\varepsilon$ with $\hat\varepsilon\ge\varepsilon\ge0$ and $|f^1_p(x_i+\varepsilon) - f^1_p(x_i)|>\eta$. For ease of notation let $A(x_i)=\Delta^i\mathcal{C}_\textnormal{obst}|_{x_i\pm\tilde{\varepsilon}}\oplus B^{n-1}_{1/p}$ i.e. the above reads $|\mu^{n-1}(A(x_i))-\mu^{n-1}(A(x_i+\varepsilon))|>\eta$. This implies that $\mu^{n-1}\left(\left(A(x_i)\cup A(x_i+\varepsilon)\right) \setminus \left( A(x_i)\cap A(x_i+\varepsilon)\right)\right)>\eta$. Since $A(x_i)$ and $A(x_i+\varepsilon)$ are compact this implies the existence of an open ball $B^\textnormal{o}$ in either $A(x_i)\setminus A(x_i+\varepsilon)$ or $A(x_i+\varepsilon)\setminus A(x_i)$ where any point in $B^\textnormal{o}$ has a positive Hausdorff distance to $A(x_i)$ or $A(x_i+\varepsilon)$. So, the semi-algebraic set-valued map $M : [0,1]\rightrightarrows \mathbb{R}^{n-1}, M(x_i)=\Delta^i\mathcal{C}_\textnormal{obst}|_{x_i\pm\tilde{\varepsilon}}\oplus B^{n-1}_{1/p}$ cannot be Hausdorff continuous in $x_i$ from the left.

Accordingly for $M$ continuous in $x_i$ from the right.
\eproof

\blemma
\label{Lem:woball_cont_ball_cont}
If the map $[0,1]\rightrightarrows \mathbb{R}^{n-1}: M(x_i)=\Delta^i\mathcal{C}_\textnormal{obst}|_{x_i\pm\tilde{\varepsilon}}$ is continuous from the left (right), then also the map $[0,1]\rightarrow \mathbb{R}^{n-1}: M'(x_i)=\Delta^i\mathcal{C}_\textnormal{obst}|_{x_i\pm\tilde{\varepsilon}}\oplus B^{n-1}_{1/p}$ is  continuous from the left (right).
\elemma
\bproof
From $M(x_i)=\Delta^i\mathcal{C}_\textnormal{obst}|_{x_i\pm\tilde{\varepsilon}}$ is continuous from the left (right) in $x_i$ it follows that $\forall\eta>0 \ \ \exists\hat{\varepsilon}>0 \ \ \forall\hat{\varepsilon}>\varepsilon>0$ it holds  $d_H(M(x_i-\varepsilon),M(x_i))<\eta$. That means $\forall q\in M(x_i-\varepsilon) \ \ \exists q'\in M(x_i)$ with $q\in q'\oplus B_\eta$ and $\forall q'\in M(x_i) \ \ \exists q\in M(x_i-\varepsilon)$ with $q'\in q\oplus B_\eta$. But then also it holds $\forall q''\in B^{n-1}_{1/p}$ that $\forall q\in M(x_i-\varepsilon) \ \ \exists q'\in M(x_i)$ with $q+q''\in q'+q''\oplus B_\eta$ and $\forall q'\in M(x_i) \ \ \exists q\in M(x_i-\varepsilon)$ with $q'+q''\in q+q''\oplus B_\eta$ and hence $M'$ is continuous in $x_i$ from the left. Accordingly for continuity from the right.
\eproof

\blemma
\label{Lem:woball_finite_disc}
$M : [0,1]\rightrightarrows \mathbb{R}^{n-1}, M(x_i)=\Delta^i\mathcal{C}_\textnormal{obst}|_{x_i\pm\tilde{\varepsilon}}$ is discontinuous in finitely many $x_i\in [0,1]$ only.
\elemma
\bproof
As a corollary of their main result Daniilidis and Pang \cite{daniilidis2011} show that:

\emph{ A closed-valued semi-algebraic set-valued map $S : \mathcal{X} \rightrightarrows \mathbb{R}^m$, where $\mathcal{X} \subset \mathbb{R}^n$ is semi-algebraic, is continuous and set-valued differentiable outside a set of dimension at most $(\dim \mathcal{X} - 1)$.}

For our case we take $\mathcal{X}=[0,1] \subset \mathbb{R}^1$ which is semi-algebraic and conclude that the closed-valued semi-algebraic set-valued map $M : \mathcal{X} \rightrightarrows \mathbb{R}^{n-1}, M(x_i) =  \Delta^i\mathcal{C}_\textnormal{obst}|_{x_i\pm\tilde{\varepsilon}}$ is discontinuous on a set of dimension at most $(\dim \mathcal{X} - 1)=0$. Thus, it can be discontinuous only on a point set $\mathcal{D}\subset [0,1]$. For the 1-dimensional case, from the derivation of this result using a finite Whitney stratification with finitely many critical values we can even conclude that $\mathcal{D}$ is finite. Thus, $M : [0,1]\rightrightarrows \mathbb{R}^{n-1}, M(x_i)=\Delta^i\mathcal{C}_\textnormal{obst}|_{x_i\pm\tilde{\varepsilon}}$ is discontinuous in finitely many $x_i\in [0,1]$ only.
\eproof

Combining the results of Lemmas \ref{Lem:ball_cont_mu_cont} -- \ref{Lem:woball_finite_disc} we get
\blemma
\label{Lem:mu_finite_disc}
The function $f^1_p(x_i)=\mu^{n-1}(\Delta^i\mathcal{C}_\textnormal{obst}|_{x_i\pm\tilde{\varepsilon}}\oplus B^{n-1}_{1/p})$ is discontinuous in finitely many $x^\textnormal{d}_{i}\in[0,1]$ only.
\elemma
\bproof
Lemma \ref{Lem:woball_finite_disc} states that the set-valued map $M(x_i)=\Delta^i\mathcal{C}_\textnormal{obst}|_{x_i\pm\tilde{\varepsilon}}$ is discontinuous in finitely many $x^\textnormal{d}_i\in [0,1]$ only. From Lemma~\ref{Lem:woball_cont_ball_cont} we get that $M'(x_i)=\Delta^i\mathcal{C}_\textnormal{obst}|_{x_i\pm\tilde{\varepsilon}}\oplus B^{n-1}_{1/p}$ can be discontinuous only in $x_i$ where $M(x_i)=\Delta^i\mathcal{C}_\textnormal{obst}|_{x_i\pm\tilde{\varepsilon}}$ is discontinuous and from Lemma~\ref{Lem:ball_cont_mu_cont} we conclude that for $f^1_p(x_i)=\mu^{n-1}(\Delta^i\mathcal{C}_\textnormal{obst}|_{x_i\pm\tilde{\varepsilon}}\oplus B^{n-1}_{1/p})$ being discontinuous in $x_i$, $M'(x_i)=\Delta^i\mathcal{C}_\textnormal{obst}|_{x_i\pm\tilde{\varepsilon}}\oplus B^{n-1}_{1/p}$ has to be discontinuous in $x_i$. Thus, $f^1_p(x_i)=\mu^{n-1}(\Delta^i\mathcal{C}_\textnormal{obst}|_{x_i\pm\tilde{\varepsilon}}\oplus B^{n-1}_{1/p})$ can be discontinuous in finitely many $x^\textnormal{d}_{i}\in[0,1]$ only.
\eproof

\blemma
\label{Lem:disc_le_eps}
If $f_p^1(x_i)$ is discontinuous in $x^\textnormal{d}_i$ from the left, then $f_p^1(x^\textnormal{d}_i)>f_p^1(x^\textnormal{d}_i-\varepsilon)$ for sufficiently small $\varepsilon$ and $f_p^1(x^\textnormal{d}_i)>f_p^1(x^\textnormal{d}_i+\varepsilon)$ for $f_p^1$ discontinuous in $x^\textnormal{d}_i$ from the right.
\elemma
\bproof
As shown in the preceding Lemma, if $f^1_p$ is discontinuous in $x_i^\textnormal{d}$ from the left (right), then also the set-valued map $M(x_i)=\Delta^i\mathcal{C}_\textnormal{obst}|_{x_i\pm\tilde{\varepsilon}}$ has to be Hausdorff discontinuous in $x_i^\textnormal{d}$ from the left (right).

Assume $M(x_i)$ is Hausdorff discontinuous from the left. Then $\exists\eta\ \ \forall\hat\varepsilon>0\ \ \exists\varepsilon$ with $\hat\varepsilon\ge\varepsilon\ge0 \ \ \exists q\in M(x_i^\textnormal{d})$ with $q\notin M(x_i^\textnormal{d}-\varepsilon) \oplus B_\eta$ or $\exists q'\in M(x_i^\textnormal{d}-\varepsilon)$ with $q'\notin M(x_i^\textnormal{d}) \oplus B_\eta$. Due to the compactness of $\Delta^i\mathcal{C}_\textnormal{obst}|_{x_i\pm\tilde{\varepsilon}}$ there cannot be an $q'\in M(x_i^\textnormal{d}-\varepsilon)$ with $q'\notin M(x_i^\textnormal{d}) \oplus B_\eta$ for all $\varepsilon>0$. Hence, $\exists\eta\ \ \forall\hat\varepsilon>0\ \ \exists\varepsilon$ with $\hat\varepsilon\ge\varepsilon\ge0 \ \ \exists q\in M(x_i^\textnormal{d})$ with $q\notin M(x_i^\textnormal{d}-\varepsilon) \oplus B_\eta$. But then, $\lim_{\varepsilon\rightarrow0}M(x_i^\textnormal{d}-\varepsilon) \subsetneq M(x_i^\textnormal{d})$ and $\lim_{\varepsilon\rightarrow0}M(x_i^\textnormal{d}-\varepsilon)\oplus B_{1/p} \subseteq M(x_i^\textnormal{d}) \oplus B_{1/p}$ and hence $\mu^{n-1}(\lim_{\varepsilon\rightarrow0}M(x_i^\textnormal{d}-\varepsilon)\oplus B_{1/p}) \le \mu^{n-1}(M(x_i^\textnormal{d}) \oplus B_{1/p})$. So if $f^1_p$ is discontinuous in $x_i^\textnormal{d}$ from the left, then $f^1_p(x_i^\textnormal{d})>f^1_p(x_i^\textnormal{d}-\varepsilon)$ for $\varepsilon$ small enough.

Accordingly for $f^1_p$ discontinuous from the right.
\eproof

\blemma
\label{Lem:f1_on_interval}
Let $I_k = [x^\textnormal{d}_k, x^\textnormal{d}_{k+1}]$ with $x^\textnormal{d}_0=0$ and $x^\textnormal{d}_{k^\textnormal{d}+1}=1$ and $x^\textnormal{d}_k$ for $1\le k\le k^\textnormal{d}$ the $k^\textnormal{d}$ points in $[0,1]$ where $f^1_p$ is discontinuous as required in Lemma~\ref{Lem:f1_converges_uniformly}. For each $k\in [0,\dots,k^\textnormal{d}]$ let now $\hat{f}^1_{p,k}:I_k\rightarrow\mathbb{R}, \hat{f}^1_{p,k}(x_i)=f^1_p(x_i)$ for $x_i\in(x^\textnormal{d}_k,x^\textnormal{d}_{k+1})$ and $\hat{f}^1_{p,k}(x^\textnormal{d}_k)=\lim_{x_i'\searrow {x^\textnormal{d}_k}} f^1_p({x'}_i)$ and $\hat{f}^1_{p,k}(x^\textnormal{d}_{k+1})=\lim_{x_i'\nearrow {x^\textnormal{d}_{k+1}}} f^1_p({x_i}')$. Then $\hat{f}^1_{p,k}$ converges uniformly to an $\hat{F}^1_k$ with $\forall x_i\in\textnormal{int}(I_k),\ \hat{F}^1_k(x_i)=F^1(x_i)$.
\elemma
\bproof
We first show that the limit from the right in $x_k^\textnormal{d}$,\  $\lim_{x_i'\searrow {x^\textnormal{d}_k}} f^1_p({x'}_i)$ and the limit from the left in $x_{k+1}^\textnormal{d}$,\ $\lim_{x_i'\nearrow {x^\textnormal{d}_{k+1}}} f^1_p({x_i}')$ exist. $f^1_p$ is discontinuous in $x_k^\textnormal{d}$. Thus it is discontinuous in $x_k^\textnormal{d}$ from the left and / or from the right. If $f^1_p$ is continuous in $x_k^\textnormal{d}$ from the right then trivially the limit $\lim_{x_i'\searrow {x^\textnormal{d}_k}} f^1_p({x'}_i)$ exists and is equal to $f^1_p(x_k^\textnormal{d})$. Accordingly for the limit from the left in $x_k^\textnormal{d}$ if $f^1_p$ is continuous from the left in $x_{k+1}^\textnormal{d}$.

If $f^1_p$ is discontinuous in $x_k^\textnormal{d}$ from the right then
\begin{align*}
\lim_{x_i'\searrow {x^\textnormal{d}_{k}}} f^1_p({x_i}')&=\lim_{x_i'\searrow {x^\textnormal{d}_{k}}} \mu^{n-1}(\Delta^i\mathcal{C}_\textnormal{obst}|_{x_i'\pm\tilde{\varepsilon}}\oplus B^{n-1}_{1/p}).
\end{align*}

For all $p,k$, $\hat{f}^1_{p,k}(x_k^\textnormal{d}) = f^1_{p,k}(x_k^\textnormal{d})$ on $(x_k^\textnormal{d}, x_{k+1}^\textnormal{d})$. Thus, from the continuity of $f^1_p$ on the open interval we directly conclude continuity of $\hat{f}^1_{p,k}$ on the open interval. But since $\hat{f}^1_{p,k}(x^\textnormal{d}_k)=\lim_{x'\searrow {x^\textnormal{d}_k}} f^1_p({x'}_i)$ and $\hat{f}^1_{p,k}(x^\textnormal{d}_{k+1})=\lim_{x'\nearrow {x^\textnormal{d}_{k+1}}} f^1_p({x_i}')$, $\hat{f}^1_{p,k}$ is continuous from the right in $x^\textnormal{d}_k$ and continuous from the left in $x^\textnormal{d}_{k+1}$ but then $\hat{f}^1_{p,k}$ is continuous also on the closed interval. From the definition of $f^1_p$ and hence $\hat{f}^1_p$ we get $\hat{f}^1_{p+1}\le\hat{f}^1_p$.

But now we have a decreasing sequence of continuous functions $\hat{f}^1_{p,k}$ converging pointwise on a compact interval $I_k$ to a continuous function $\hat{F}^1_k$. Dini's theorem gives uniform convergence on $I_k$.
\eproof

\bproof[Proof of Lemma~\ref{Lem:uniform_convergence}]
But now we have shown that both expressions required in Lemma~\ref{Lem:uniform_convergence_if} converge uniformly and hence also $$\mu^{n-1}(\Delta^i\mathcal{C}_\textnormal{obst}|_{x_i\pm\varepsilon}\oplus B^{n-1}_\varepsilon)\xrightarrow[\varepsilon\rightarrow0]{\textnormal{uniform}}0.$$
\eproof

Now we have bounded the probability of finding a \emph{good} sample from below independent of $\kappa$ and $\varepsilon$ and have shown that by shrinking the pruned $\varepsilon$-balls we can let the upper bound on finding a \emph{bad} sample go to zero for any inner or outer cell of a pruned $\varepsilon$-ball containing a straight line segment of $\gamma$ but not intersecting with a hyperplane $H_{x_i}$

Thus, we can choose $\varepsilon$ such that whenever an inner cell of a pruned $\varepsilon$-ball of the finite covering of $\gamma$ is sampled, the probability of finding a \emph{good} sample clearing the entire intersection of $\kappa$ with the pruned $\varepsilon$-ball is larger than the probability of finding a \emph{bad} sample potentially leading to a cell split in traversing dimension.

What is left to show is that neither the pruned $\varepsilon$-balls holding the cells intersecting with a hyperplane $H_{x_i}$ with nonzero $\mu^{n-1}$ nor pruned $\varepsilon$-balls holding a corner of $\gamma$ can be sampled arbitrarily often.

We have shown that for any dimension $i$ there exist only a finite number of hyperplanes $H_{x_i}$ as defined above.

\blemma
\label{Lem:most_one_hyperplane}
There exists an $\hat{\varepsilon}>0$ such that for any $\varepsilon<\hat\varepsilon$ any pruned $\varepsilon$-ball $\overline{B}_{\varepsilon}$ with traversing dimension $i$ intersects with at most one hyperplane $H_{x_i}$.
\elemma
\bproof
Let $X_i=\{x_i|\mu^{n-1}({\delta\mathcal{C}_\textnormal{obst}}_{|_{x_i}})>0\}$ be the set of $x_i$ where corresponding hyperplanes exist. Then $\min_{x_i',x_i''\in X_i}d({x_i',x_i''})$ is the minimum distance in dimension $i$ between any two of these hyperplanes. Thus, with $\hat\varepsilon = \min_{x_i',x_i''\in X_i}d({x_i',x_i''}) / 2$ the Lemma holds, since such a ball would have a width of less than the minimum distance of any two hyperplanes $H_{x_i}$.
\eproof
So for any inner cell of such a pruned $\varepsilon$-ball with traversing dimension $i$ we can be sure that it intersects with at most one hyperplane $H_{x_i}$. For the outer cells of such a pruned $\varepsilon$-ball this is unfortunately not true and we cannot affect the number of hyperplanes an outer cell intersects with by choosing the size of the ball. Fortunately this is not necessary. Recall the derivation of the upper bound of finding a bad sample in an outer cell in Lemma~\ref{Lem:bad_bound_outer}. It showed that the probability of finding a sample leading to a split in traversing dimension $i$ depends on $\delta\mathcal{C}_\textnormal{obst}$ in a volume that is bounded in dimension $i$ by $c^{\overline{B}_\varepsilon\textnormal{u}}_i\pm D^I_i$ where $D^I_i$ is the width of the intersection of the cell and the ball. Thus, a hyperplane $H_{x_i}$ might affect the probability of finding a \emph{bad} sample in an outer cell of a pruned $\varepsilon$-ball $\overline{B}_\varepsilon$ only if $x_i \in \left[c^{\overline{B}_\varepsilon\textnormal{l}}_i-2\varepsilon, c^{\overline{B}_\varepsilon\textnormal{u}}_i+2\varepsilon\right]$.
\blemma
\label{Lem:only_one_h}
There exists an $\hat{\varepsilon}>0$ such that for any $\varepsilon<\hat\varepsilon$ the probability of finding a \emph{bad} sample in any pruned $\varepsilon$-ball $\overline{B}_{\varepsilon}$ with traversing dimension $i$ is affected by at most one hyperplane $H_{x_i}$.
\elemma
\bproof
Thus, we have to ensure, that in dimension $i$ at most one hyperplane lies inside $\left[c^{\overline{B}_\varepsilon\textnormal{l}}_i-2\varepsilon, c^{\overline{B}_\varepsilon\textnormal{u}}_i+2\varepsilon\right]$. But $d\left(c^{\overline{B}_{\varepsilon}\textnormal{l}}_i, c^{\overline{B}_{\varepsilon}\textnormal{u}}_i\right) = 2 \varepsilon$, so any two hyperplanes must be at least $6\varepsilon$ apart.
Thus, with $\hat{\varepsilon} = \min_{x_i',x_i''\in X_i}d({x_i',x_i''}) / 6$ as of Lemma~\ref{Lem:most_one_hyperplane} the Lemma holds.
\eproof

So we have shown that in any pruned $\varepsilon$-ball there is at most one cell where the probability of finding a \emph{bad} sample splitting the cell in traversing dimension $i$ inside the ball is affected by a hyperplane $H_{x_i}$. We denote this cell with $\kappa^\textnormal{hyp}$. Whenever $\kappa^\textnormal{hyp}$ is split in the traversing dimension $i$, only one of the resulting new cells intersects with $H_{x_i}$. The label $\kappa^\textnormal{hyp}$ is passed to this new cell intersecting with $H_{x_i}$.

\blemma
\label{Lem:hyp_prob_finite}
For a cell $\kappa^\textnormal{hyp}$ intersecting with a pruned $\varepsilon$-ball and a hyperplane $H_{x_i}$ the expected number of times the respective cell intersecting with $H_{x_i}$ gets split is bounded from above.
\elemma
\bproof
Of course also for any cell holding the label $\kappa^\textnormal{hyp}$ the general lower bound on the probability of finding a \emph{good} sample holds and is equal to $\varepsilon^n$. But then, the probability of finding a \emph{good} sample after at most $k$ splits is larger or equal to the probability of \emph{success} after at most $k$ Bernoulli trials with $p = \varepsilon^n$ and we can derive an upper bound on the expected value of times $\kappa^\textnormal{hyp}$ gets split before being cleared around $\gamma$ from the geometric distribution being $E = 1/p$. Thus, the expected value of times $\kappa^\textnormal{hyp}$ is split before it is cleared around $\gamma$ is bounded by $E\le\varepsilon^{-n}$ and hence finite.
\eproof

So we have shown that any pruned $\varepsilon$-ball holding a straight line segment of $\gamma$ with $\varepsilon$ small enough according to above lemmata will not be sampled arbitrarily often but will be cleared after a bounded expected number of times it got sampled.

\msubsubsection{Even pruned $\varepsilon$-balls holding a corner of $\gamma_\textnormal{Man}$ will not be sampled arbitrarily often}

According to Lemma~\ref{Lem:straight_or_corner}, for $\varepsilon<\tilde{\varepsilon}$ any pruned $\varepsilon$-ball around $\gamma$ holds either a straight line segment or a single corner connecting two straight line segments of $\gamma$. We have shown so far that pruned $\varepsilon$-balls holding a straight line segment of the path will not be sampled arbitrarily often but will be cleared before. We will now extend this result to pruned $\varepsilon$-balls holding a corner configuration. Let $\overline{B}^\textnormal{cor}_\varepsilon$ be a pruned $\varepsilon$-ball holding a single corner of $\gamma$.

Now for any such $\overline{B}^\textnormal{cor}_\varepsilon$ at any time there exists exactly one possibly free or possibly occupied cell $\kappa^\textnormal{cor}$ that holds the corner configuration of $\gamma$. When this cell gets split, only one of the two resulting cells will hold $\gamma(l^\textnormal{cor})$ and the label $\kappa^\textnormal{cor}$ is passed on to this cell.
We can neglect cases where this corner configuration exactly lies on the boundary of two or more cells. The exact location of the corner configuration given by the algorithm in the proof of Lemma~\ref{Lem:manhattan} is arbitrary and there exists a whole continuum of Manhattan paths.

All other cells intersecting with $\overline{B}^\textnormal{cor}_\varepsilon$ contain a straight line segment of $\gamma$ only. For these cells there exists a well defined traversing dimension $i$, so we can use the results found above for $\varepsilon$-balls holding a straight line segment of the path.

For $\kappa^\textnormal{cor}$ the same lower bound on finding a \emph{good} sample holds. We actually don't care about a diminishing upper bound for a \emph{bad} sample. Instead for our aim to prove probabilistic completeness, the following Lemma is sufficient.

\blemma
For a pruned $\varepsilon$-ball $\overline{B}^\textnormal{cor}_\varepsilon$ holding a single corner configuration $\gamma(l^\textnormal{cor})$ the expected number of times the respective cell holding the corner configuration $\kappa^\textnormal{cor}$ gets split is bounded from above.
\elemma
\bproof
Of course also for any cell holding the label $\kappa^\textnormal{cor}$ the general lower bound on finding a \emph{good} sample holds and is equal to $\varepsilon^n$. Thus, the upper bound on the expected number of times $\kappa^\textnormal{cor}$ gets split before being cleared around $\gamma$ is equal to the result found in Lemma~\ref{Lem:hyp_prob_finite} and hence finite.
\eproof

Thus, a pruned $\varepsilon$-ball holding a corner connecting two straight line segments of $\gamma$ will not be sampled arbitrarily often but will be cleared after a bounded expected number of times it got sampled.

\msubsubsection{The probability that PCD solves the path planning query approaches one with the number of iterations of PCD going to infinity}
\blemma
If all pruned $\varepsilon$-balls of a finite covering of a pruned $\varepsilon$-tunnel around $\gamma_\textnormal{Man}$ are cleared, the path planning query is solved and part \texttt{<a>} and hence PCD return with \texttt{success=true}.
\elemma
\bproof
If all pruned $\varepsilon$-balls of a finite covering of a pruned $\varepsilon$-tunnel around $\gamma_\textnormal{Man}$ are cleared, all cells intersecting with any pruned $\varepsilon$-ball are cleared around $\gamma$ and hence are possibly free. But then from the strictly positive diameter of the pruned $\varepsilon$-tunnel we get that adjacent cells of $K_\gamma=\left\{\kappa|\kappa\cap\gamma_\textnormal{Man}\neq\emptyset\right\}$ share a common area on the dividing hyperplane with nonzero $\mu^{n-1}$ and --- put in the right order --- form a cell path $\phi$. In the next iteration of \texttt{<a>}, \texttt{findCellPath} will consequentially return a cell path. A continuous path through this channel is then checked for collision. If any collisions are found, cells along the cell path get split. However, non of the splits will affect any of the pruned $\varepsilon$-balls since these are cleared and will never again be contained in a possibly occupied cell. Hence, in all remaining iterations of \texttt{<a>} \texttt{findCellPath} will return a cell path.
However, the result of Lemma~\ref{Lem:a_returns} limits the number of iterations of \texttt{<a>} after which the \texttt{while} loop breaks. Since still a cell path can be found, 
\texttt{checkPath(cellPath)} will return \texttt{true} and consequentially \texttt{<a>} and PCD will return with \texttt{success=true}.
\eproof

\bproof{Proof of Proposition~\ref{Prop:completeness}}

But now we have shown that
\begin{itemize}
  \item $\gamma$ can be assumed to be of Manhattan-type.
  \item $\varepsilon$ can be chosen small enough such that the $\varepsilon$-tunnel around $\gamma$ is collision free with a finite covering by $\varepsilon$-balls.
  \item In each iteration of PCD \texttt{<a>} returns after a bounded number of iterations and hence the total number of iterations of PCD goes to infinity.
  \item In each iteration of PCD the probability of finding a sample in any $\varepsilon$-tunnel can be bounded from below independent of $\varepsilon$ and hence in at least one ball of the finite covering, the number of samples goes to infinity with the number of iterations of PCD.
  \item $\varepsilon$ can be chosen small enough such that the expected number of times a pruned $\varepsilon$-ball gets sampled before it is cleared is finite for both $\varepsilon$-balls holding a corner and $\varepsilon$-balls holding a straight line segment of the path.
  \item At the latest when all pruned $\varepsilon$-balls of the finite covering are cleared, PCD returns success.
  \item Consequentially, the expected number of iterations after which PCD returns success is bounded and hence PCD is probabilistically complete.
\end{itemize}
\eproof

\section{Conclusion and Future Work}
We have shown that PCD is probabilistically complete for  a
configuration space $\mathcal{C}=[0, 1]^n$ with $\mathcal{C}$-obstacles
modeled as semi-algebraic sets. 

This proof is based on the assumption that collision-free samples found while checking a continuous path in part \texttt{<a>} of Algorithm \ref{Fig:pcdbalg} are not stored in the respective cell. By not using these samples, we through away valuable and often costly information about the shape of $\mathcal{C}_\textnormal{free}$. Even being crucial to some parts of the proof, we are confident that incorporating these samples does not harm probabilistic completeness. From the experience of using PCD on a variety of path planning problems, we have no evidence that PCD could be not probabilistically complete when doing so. Therefore, for applications we propose to continue to store these samples in the collision-free cells. It remains to show that PCD is still probabilistically complete when incorporating these samples.

\section*{Acknowledgment}
\bibliographystyle{IEEEtran}
\bibliography{IEEEabrv,../../../Thesis/mybibref}

\end{document}